\documentclass[conference]{IEEEtran}

\IEEEoverridecommandlockouts
\usepackage{cite}
\usepackage{amsmath,amssymb,amsfonts}
\usepackage{algorithm}
\usepackage{algorithmic}
\usepackage{graphicx}
\usepackage{textcomp}
\usepackage{xcolor}
\usepackage{tabularx}
\usepackage{multirow}
\usepackage{multicol}
\usepackage{booktabs}
\def\BibTeX{{\rm B\kern-.05em{\sc i\kern-.025em b}\kern-.08em
    T\kern-.1667em\lower.7ex\hbox{E}\kern-.125emX}}
    
\title{A Survey of Deep Learning Techniques for Neural Machine Translation}

\author{\IEEEauthorblockN{Shuoheng Yang, Yuxin Wang, Xiaowen Chu}
\IEEEauthorblockA{\textit{Department of Computer science } \\
\textit{Hong Kong Baptist University}\\
Hong Kong, China \\
yshuoheng@gmail.com, \{yxwang, chxw\}@comp.hkbu.edu.hk}
}
\begin{document}

\maketitle

\begin{abstract}
In recent years, natural language processing (NLP) has got great development with deep learning techniques. In the sub-field of machine translation, a new approach named Neural Machine Translation (NMT) has emerged and got massive attention from both academia and industry. However, with a significant number of researches proposed in the past several years, there is little work in investigating the development process of this new technology trend. This literature survey traces back the origin and principal development timeline of NMT, investigates the important branches, categorizes different research orientations, and discusses some future research trends in this field.
\end{abstract}


\section{Introduction}

\subsection{Introduction of Machine Translation}

Machine translation (MT) is a classic sub-field in NLP that investigates how to use computer software to translate the text or speech from one language to another without human involvement. Since MT task has a similar objective with the final target of NLP and AI, i.e., to fully understand the human text (speech) at semantic level, it has received great attention in recent years. Besides the scientific value, MT also has huge potential of saving labor cost in many practical applications, such as scholarly communication and international business negotiation.

Machine translation task has a long research history with many efficient methods proposed in the past decades. Recently, with the development of Deep Learning, a new kind of method called Neural Machine Translation (NMT) has emerged. Compared with the conventional method like Phrase-Based Statistical Machine Translation (PBSMT), NMT takes advantages in its simple architecture and ability in capturing long dependency in the sentence, which indicates a huge potential in becoming a new trend of the mainstream. After a primitive model in origin, there are a variety of NMT models being proposed, some of which have achieved great progresses with the state-of-the-art result. 
This paper summarizes the major branches and recent progresses in NMT and discusses the future trend in this field.

\subsection{Related Work and Our Contribution}

Although there is little work in the literature survey of NMT, some other works are highly related. Lipton et al. have summarized the conventional methods in sequence learning\cite{Lipton2015}, which provided essential information for the origin of NMT as well as the related base knowledge. Britz et al. and Tobias Domhan have done some model comparison work in NMT with experiment and evaluation in the practical performance of some wildly accepted technologies, but they have rare theoretical analysis, especially in presenting the relationship between different proposed models\cite{Britz2017}\cite{Domhan2018}. On the other hand, some researchers limited their survey work on a special part related to NMT — the Attention Mechanism, but both of them have a general scope that oriented to all kinds of AI tasks with Attention \cite{Chaudhari2019}\cite{Galassi2019}. Maybe the most related work was an earlier doctoral thesis written by Minh-Thang Luong in 2016\cite{Luong2017}, which included a comprehensive description about the original structure of NMT as well as some wildly applied tips. 

This paper, however, focuses on a direct and up-to-date literature survey about NMT. We have investigated a lot about the relevant literature in this new trend and provided comprehensive interpretation for current mainstream technology in NMT. 

As for concrete components, this literature survey investigates the origin and recent progresses in NMT, categorizes these models by their different orientation in the model structure. Then we demonstrate the insight of these NMT types, summarize the strengths and weaknesses by reviewing their design principal and corresponding performance analysis in translation quality and speed. We also give a comprehensive overview of two components in NMT development, namely attention mechanism and vocabulary coverage mechanism, both of which are indispensable for current achievement. At last, we give concentration on some literature which proposed advanced models with comparison work; we introduce these considerable models as well as the potential direction in future work.

Regarding the survey scope, some subareas of NMT with less attention were deliberately left out of the scope except with brief description in future trend. These include but are not limited to the literature of robustness of NMT, domain adaptation in NMT and other applications that embed NMT method (such as speech translation, document translation). Although the research scope has been specifically designed, due to the numerous of researches and the inevitable expert selection bias, we believe that our work is merely a snapshot of part of current research rather than all of them. We are hoping that our work could provide convenience for further research.

The remaining of the paper is organized as follows.
Section~\ref{sec:Sec2} provides an introduction of machine translation and presents its history of development. Section~\ref{sec:Sec3} introduces the structure of NMT and the procedure of training and testing. Section~\ref{sec:Sec4} discusses attention mechanism, an essential  innovation in the development of NMT. Section~\ref{sec:Sec5} surveys a variety of methods in handling word coverage problem and some fluent divisions. Section~\ref{sec:Sec6} describes three advanced models in NMT. Finally, Section~\ref{sec:Sec7} discusses the future trend in this field.
\section{History of Machine Translation}
\label{sec:Sec2}
Machine translation (MT) has a long history; the origin of this field could be traced back to the 17th century. In 1629, Ren\'e Descartes came up with a universal language that expressed the same meaning in different languages and shared one symbol.

The specific research of MT began at about 1950s, when the first researcher in the field, Yehoshua Bar-Hillel, began his research at MIT (1951) and organized the first International Conference on Machine Translation in 1952. Since then, MT has experienced three primary waves in its development, the Rule-based Machine Translation\cite{Forcada2011}, the Statistical Machine Translation\cite{Koehn2003}\cite{Koehn2007}, and the Neural Machine Translation\cite{Cho2014}. We briefly review the development of these three stages in the following.

\subsection{Development of Machine Translation}

\subsubsection{Rule-based Machine Translation}

Rule-based Machine Translation is the first design in MT, which is based on the hypothesis that all different languages have its symbol in representing the same meaning. Because in usual, a word in one language could find its corresponding word in another language with the same meaning. 

In this method, the translation process could be treated as the word replacement in the source sentence. In terms of 'rule-based', since different languages could represent the same meaning of sentence in different word order, the word replacement method should base on the syntax rules of both two languages. Thus every word in the source sentence should take its corresponding position in the target language.

The rule-based method has a beautiful theory but hardly achieves satisfactory performance in implementation. This is because of the computational inefficiency in determining the adaptive rule of one sentence. Besides, grammar rules are also hard to be organized, since linguists summarize the grammar rules, and there are too many syntax rules in one language (especially language with more relaxed grammar rules). It is even possible that two syntax rules conflict with each other.

The most severe drawback of rule-based method is that it has ignored the need of context information in the translation process, which destroys the robustness of rule-based machine translation. One famous example was given by Marvin Minsky in 1966, where he used two sentences given below:
 
``$The$ $pen$ $is$ $in$ $the$ $box$''

``$The$ $box$ $is$ $in$ $the$ $pen$''

Both sentences have the same syntax structure. The first sentence is easy to understand; but the second one is more confusing, since the word ``pen'' is a polysemant, which also means ``fence'' in English. But it is difficult for the computer to translate the ``pen'' to that meaning; the word replacement is thus an unsuccessful method.

\subsubsection{Statistical Machine Translation}
Statistical Machine Translation (SMT) has been the mainstream technology for the past 20 years. It has been successfully applied in the industry, including Google translation, Baidu translation, etc.

Different from Rule-based machine translation, SMT tackles the translation task from a statistical perspective. Concretely, the SMT model finds the words (or phrases) which have the same meaning through bilingual corpus by statistics. Given one sentence, SMT divides it into several sub-sentences, then every part could be replaced by target word or phrase.

The most prevalent version of SMT is Phrase-based SMT (PBSMT), which in general includes pre-processing, sentence alignment, word alignment, phrase extraction, phrase feature preparation, and language model training. The key component of a PBSMT model is a phrase-based lexicon, which pairs phrases in the source language with phrases in the target language. The lexicon is built from the training data set which is a bilingual corpus. By using phrases in this translation, the translation model could utilize the context information within phrases. Thus PBSMT could outperform the simple word-to-word translation methods.

\subsubsection{Neural Machine Translation}
It has been a long time since the first try on MT task by neural network\cite{Allen1987}\cite{Chrisman1991}. Because of the poor performance in the early period and the computing hardware limitation, related research in translation by neural network has been ignored for many years. 

Due to the proliferation of Deep Learning in 2010, more and more NLP tasks have achieved great improvement. Using deep neural networks for MT task has received great attention as well. A successful DNN based Machine Translation (NMT) model was first proposed by Kalchbrenner and Blunsom\cite{Kalchbrenner2013}, which is a totally new concept for MT by that time. Comparing with other models, the NMT model needs less linguistic knowledge but can produce a competitive performance. Since then, many researchers have reported that NMT can perform much better than the traditional SMT model\cite{Klein2017}\cite{Bojar2015}\cite{Cettolo2015}\cite{Junczys-Dowmunt2016}\cite{Bentivogli2016}, and it has also been massively applied to the industrial field\cite{Wu2016}. 


\begin{figure}[b]
\centerline{\includegraphics[width=18pc]{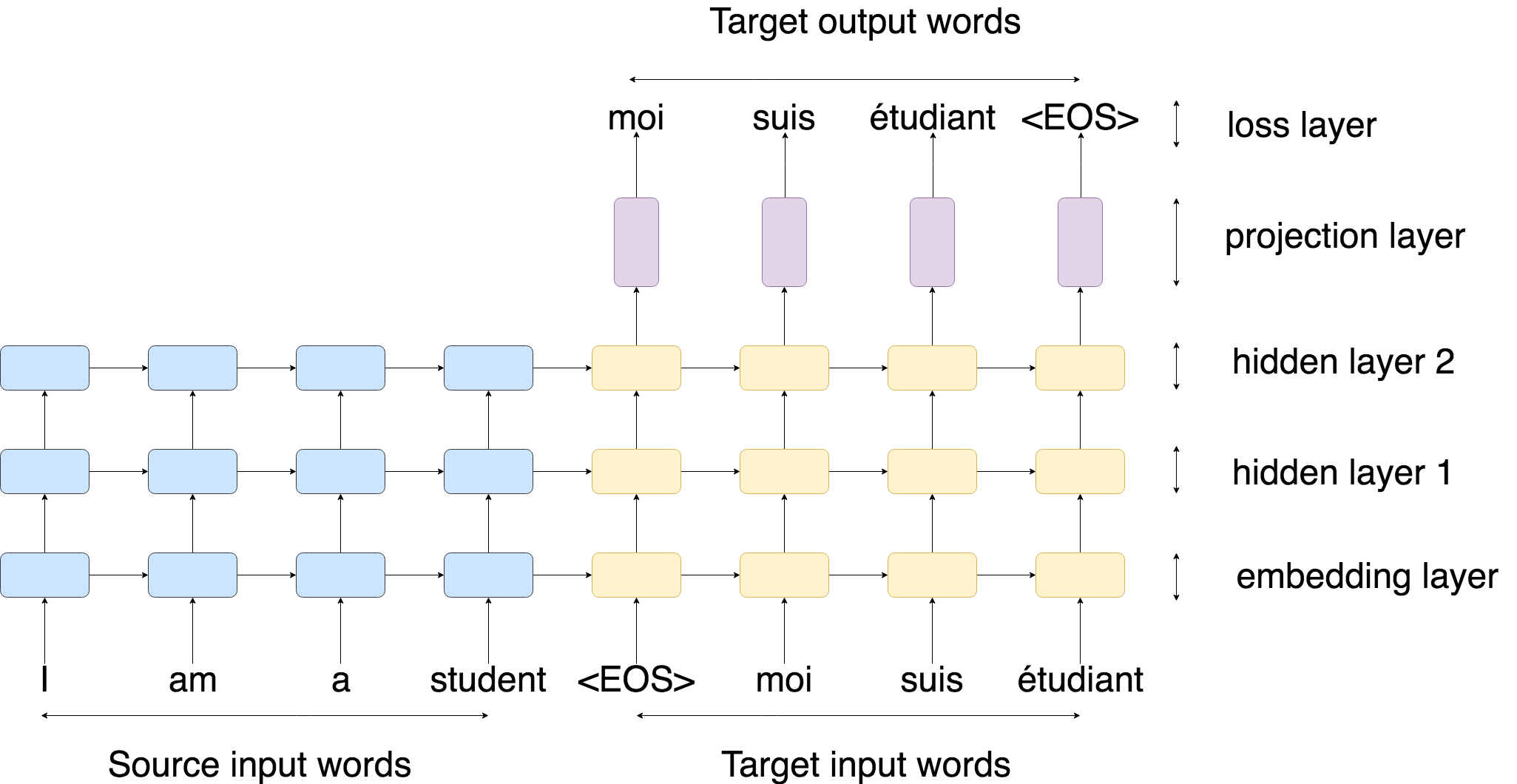}}
\caption{The training process of RNN based NMT. The symbol $<EOS>$ means end of sequence. The embedding layer is for pre-processing. The two RNN layers are used to represent the sequence.}
%
\label{fig_env2}
\end{figure}

\subsection{Introduction of Neural Machine Translation}

\subsubsection{Motivation of NMT}

The inspiration for neural machine translation comes from two aspects: the success of Deep Learning in other NLP tasks as we mentioned, and the unresolved problems in the development of MT itself.

For the first reason, in many NLP tasks, traditional Machine Learning method is highly dependent on hand-crafted features that often come from linguistic intuition, which is definitely an empirical trial-and-error process\cite{Collobert2011}\cite{Young2018} and is often far more incomplete in representing the nature of original data. For example, the context size of training language model is assigned by researchers with strong assumption in context relation\cite{Song1999}; and in text representation method, the classic bag-of-words (BOW) method has ignored the influence of word order\cite{Wallach2006}. However, when applying deep neural network (DNN) in the aforementioned tasks, the DNN requires minimum domain-knowledge and avoids some pre-processing steps in human feature engineering \cite{Britz2017}.

DNN is a powerful neural network which has achieved excellent performance in many complex learning tasks which are traditionally considered difficult\cite{Krizhevsky2012}\cite{Maas2013}. In NLP field, DNN has been applied in some traditional tasks, for example, speech recognition \cite{Chorowsk2014} and Named Entity Recognition (NER) \cite{Collobert2011}. With the exceptional performance they got, DNN-based models have found many potential applications in other NLP tasks.

For the second reason, in MT field, PBSMT has got a pretty good performance in the past decades, but there are still some inherent weaknesses which require further improvement. First, since the PBSMT generates the translation by segmenting the source sentence into several phrases and doing phrase replacement, it may ignore the long dependency beyond the length of phrases and thus cause inconsistency in translation results such as incorrect gender agreements. Second, there are generally many intricate sub-components in current systems \cite{Galley2008}\cite{Chiang2009}\cite{Green2013}, e.g., language model, reordering model, length/unknown penalties, etc. With the increasing number of these sub-components, it is hard to fine-tune and combine each other to get a more stable result \cite{Luong2015}.

All the above discussions have indicated the bottleneck in the development of SMT miniature. Specifically, this bottleneck mainly comes from the language model (LM). This is because, in MT task, language model actually can give the most important information: the emergence probability of a particular word (or phrase) that is conditioned on previous words. So building a better LM can definitely improve the translation performance.

The vast majority of conventional LM is based on the Markov assumption:
\begin{equation}
\begin{aligned} p\left(x_{1}, x_{2}, \ldots, x_{T}\right) &=\prod_{t=1}^{T} p\left(x_{t} | x_{1}, \ldots, x_{t-1}\right) \\ & \approx \prod_{t=1}^{T} p\left(x_{t} | x_{t-n}, \ldots, x_{t-1}\right) \end{aligned}
\end{equation}
where $x_1, x_2, ..., x_T$ is a sequence of words in a sentence and $T$ represents the length of the sentence. 

In this assumption, the probability of the sentence is equal to the multiplication of probability of each word. $n$ is the total number of words that is chosen to simplify the model, which is also referred to as $context\ window$.

Obviously, the dependency of words that exceed $n$ would be ignored, which implies that the conventional LM performs poorly on modeling long dependency. Moreover, since the experimental result has indicated that a modest context size (generally 4-6 words) can be accepted, the first problem of traditional LM is the limited representation ability.

Besides, the data sparsity for training has always been the problem that hinders an LM built with a larger size of context window. This is because the number of $n$-tuples for counting is exponential in $n$. In other words, when building an LM, with the increment of the number of order, the number of training samples we need would also increase remarkably, which is also referred to as ''curse of dimensionality". For example, if one LM has the order of 5 with a vocabulary size of 10,000, then the possible combination of words for statistics should be about $10^{25}$, which requires enormous training data. And since most of these combinations have not been observed before, subsequent researches have used various trade-off and smoothing method to alleviate the sparsity problem\cite{Teh2006}\cite{Stolcke2002}\cite{Rosenfeld2000}\cite{Federico2008}\cite{Heafield2011}.

While further research of the aforementioned LM with statistical method has become almost stagnant, Neural Language Model (NLM)\cite{Bengio2003}, on the other hand, uses a neural network to build a language model that models text data directly. In initial stage, NLM used fixed-length of a feature vector to represent each word, and then the solid number of word vectors would concatenate together as a semantic metric to represent the context \cite{Bengio2003} \cite{Schwenk Gauvain2005}\cite{Schwenk2007}, which is very similar to the $context\ window$. This work was enhanced later by injecting additional context information from source sentence\cite{Devlin2014}\cite{Son2012}\cite{Schwenk2012}. Comparing with the traditional LM, the original NLM alleviates the sample sparsity due to the distributed representation of the word, which enables them to share the statistical weights rather than being independent variables. And since words with similar meaning may occur in the similar context, the corresponding feature vector would have the similar value, which indicates that the semantic relation of words has been ''embedded" into the feature vector.  

New proposals in the next stage solve the long dependency problem by using Recurrent Neural Network (RNN). RNN based NLM (RNLM) models the whole sentence by reading each word once a time-step, thus it can model the true conditional probability without limitation of context size\cite{Mikolov2010}. Before the emergence of NMT, the RNLM, as mentioned earlier, outperformed the conventional LM in the evaluation of text perplexity and brought better performance in many practical tasks\cite{Mikolov2010}\cite{Chung2014}.

The direct application of NLM in SMT has been naturally proposed \cite{Devlin2014}\cite{Auli2013}\cite{Auli2014}\cite{Schwenk2006}\cite{Vaswani2013}, and the preliminary experiment indicated promising results. The potential of NLM motivates further exploration for a complete DNN based translation model. Subsequently, a more ''pure" model with the only neural network has emerged, with the DNN architecture that learns to do the translation task end-to-end. Section~\ref{sec:Sec3} demonstrates its basic structure (in Fig.~\ref{fig_env3}), as well as its concrete details.

\subsubsection{Formulation of NMT Task}
\begin{figure}[h]
\centering
\includegraphics[width=18pc]{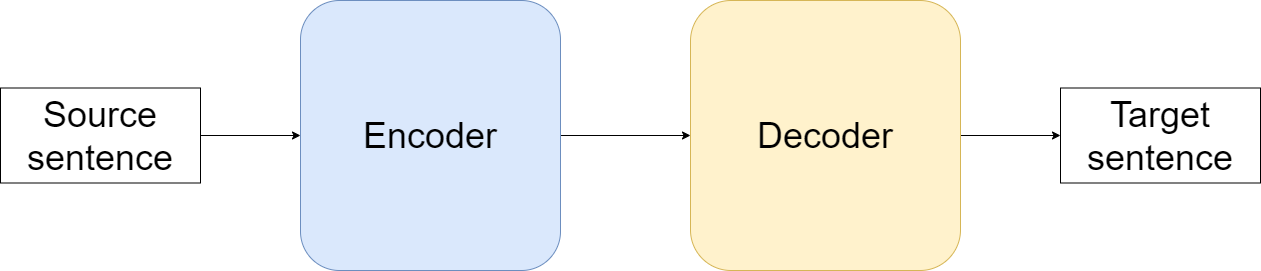}
\caption{End-to-End structure in modern NMT model. The encoder is used to represent the source sentence to semantic vector, while the decoder makes prediction from this semantic vector to a target sentence. End-to-End means the model processes source data to target data directly, without explicable intermediate result.}
%
\label{fig_env0}
\end{figure}

Currently, NMT task is originally designed as an end-to-end learning task. It directly processes a source sequence to a target sequence. The learning objective is to find the correct target sequence given the source sequence, which can be seen as a high dimensional classification problem that tries to map the two sentences in the semantic space. In all mainstreams of modern NMT model, this process can be divided into two steps: encoding and decoding, and thus can functionally separate the whole model as Encoder and Decoder as illustrated in Fig.~\ref{fig_env0}.

In perspective of probability, NMT generates the target sequence $T(t_1,t_2,...,t_m)$ from the max conditional probability given the source sequence $S(s_1,s_2,...,s_n)$, where $n$ is the length of sequence $S$ and $m$ is the length of target sequence $T$. The whole task could be formulated as\cite{Wu2016}:
\begin{equation}
 arg max\ P(T|S).
\label{eq1}
\end{equation}%
More concretely, when generating each word of the target sentence, it uses the information from both the word it predicted previously and the source sentence. In that case, each generating step could be described as when generating the $i$-th word: 
\begin{equation}
arg max\prod\limits_{i=1}^m P(t_i|t_{j<i},S)
\label{eq2}
\end{equation}%
Based on this formula and the discussion of NLM above, NMT task could be regarded as an NLM model with additional constraints (e.g., conditioned on a given source sequence).

\subsection{The Recent Development in NMT}

We devide the recent developement of NMT in five main stages: (a) the original NMT with a shallow layer, (b) SMT assisted by NLM, (c) the DNN based NMT, (d) NMT with attention mechanism, (e) the attention-based NMT. 

\textbf{NMT with Shallow Layer}

Even before the Deep Learning, Allen has used binary encoding to train an NMT model in 1987\cite{Allen1987}. Later in 1991, Chrisman used Dual-ported RAAM architecture\cite{Pollack1990} to build an original NMT model\cite{Chrisman1991}. Although both of them have a pretty primitive design with the limited result when looking back, their work has indicated the original idea of this field. The further related work has almost stagnated in the following decades, due to the huge progress that SMT method acquired at that period, as well as the limited computing power and data samples. 

\textbf{SMT assisted by NLM}

Based on the above discussion, NLM has revolutionized the traditional LM even before the rise of deep learning. Later on, deep RNN based NLM has been applied in the SMT system. Cho et al. proposed an SMT model along with an NLM model\cite{Cho2014b}. Although the main body is still SMT, this hybrid method provides a new direction for the emergence of a pure deep learning-based NMT.

\textbf{NMT with Deep Neural Network}

Since the traditional SMT model with NLM has got the state-of-the-art performance at that time, a pure DNN based translation approach was proposed later with an end-to-end design to model the entire MT process\cite{Kalchbrenner2013}\cite{Sutskever2014}. Using DNN based NMT could capture subtle irregularities in both two languages more efficiently \cite{Wu2016}, which is similar to the observation that DNNs often have a better performance than 'shallow' neural networks \cite{Bahdanau2014}.

\textbf{NMT with Attention Mechanism}

Although the initial DNN based NMT model has not outperformed the SMT completely, it still exhibited a huge potential for further research. When tracing back the major weakness, although one theoretical advantage of RNN is its ability in capturing the long dependency between words, in fact, the model performance would deteriorate with the increase of sentence length. This scenario is due to the limited feature representation ability in a fixed-length vector. Under the circumstances, since the original NMT has got a pretty good performance without any auxiliary, the idea of whether some variants in architecture could bring a breakthrough has led to the rise of Attention Mechanism.

Attention Mechanism was originally proposed by Bahdanau et al. as an intermediate component\cite{Bahdanau2014}, and the objective is to provide additional word alignment information in translating the long sentence. Surprisingly, NMT model has got a considerable improvement with the help of this simple method. Later on, with tremendous popularity among both academia and industry, many refinements in Attention Mechanism have emerged, and more details will be discussed in Section~\ref{sec:Sec4}.

\textbf{Fully Attention based NMT}

With the development of Attention Mechanism, fully Attention-based NMT has emerged as a great innovation in NMT history. In this new tendency, Attention mechanism has taken the dominate position in text feature extraction rather than a auxiliary component. And the representative model is Transformer \cite{Vaswani2017}, which is a fully attention-based model proposed by Vaswani et al. 

Abandoning previous framework in neither RNN nor CNN based NMT models, Transformer is a that solely based on an intensified version of Attention Mechanism called Self-Attention with feed-forward connection, which got revolutionary progress in structure with state-of-the-art performance. Specifically, the innovative attention structure is the secret sauce to gain such significant improvement. The self-attention is a powerful feature extractor which also allows to 'read' the entire sentence and model it once a time. In the perspective of model architecture, this character can be seen as a combination of advantages from both CNN and RNN, which endows it a good feature representation ability with high inference speed. More details about self-attention will be given in Section~\ref{sec:Sec4}. The architecture of Transformer will be discussed in Section~\ref{sec:Sec6}.

\section{DNN based NMT}
\label{sec:Sec3}
The emergence of DNN based NLM indicates the feasibility of building a pure DNN based translation model. The further implementation is the $de facto$ form of NMT in origin. 
This section reviews the basic concept of DNN based NMT, demonstrates a comprehensive introduction of the standard structure of the original DNN based NMT, and discusses the training and inferencing processes.

\subsection {Model Design of DNN based NMT}

There are many variations of network design for NMT, which can be categorized into $recurrent$ or $non-recurrent$ models. More specifically, this category can be traced back to the early development of NMT, when RNN and CNN based models are the most common design. Many sophisticated models proposed afterwards also belong to either CNN or RNN family. This sub-section follows the development of NMT in the early years, and demonstrates some representative models by classifying them as RNN or CNN based models.

\subsubsection {RNN based NMT}

Although in theory, any network with enough feature extraction ability could be selected to build an NMT model, in $de facto$ implementations, RNN based NMT models have taken the dominant position in NMT development, and they have achieved state-of-the-art performance. Based on the discussion in Section~\ref{sec:Sec2}, since many NLM literature used RNN to model the sequence data, this design has intuitively motivated the further work to build an RNN based NMT model. In the initial experiment, an RNN based NLM was applied as a feature extractor to compress the source sentence into a feature vector, which is also referred to as thought vector. Then a similar RNN was applied to do the 'inverse work' to find the target sentence that can match the previous thought vector in semantic space. 

The first successful RNN based NMT was proposed by Sutskever et al., who used a pure deep RNN model and got a performance that approximates the best result achieved by SMT\cite{Sutskever2014}. Further development proposed the Attention Mechanism, which improves the translation performance significantly and exceeds the best SMT model. GNMT model was an industry-level model applied in Google Translation, and it was regarded as a milestone in RNN based NMT.

Besides the above mentioned work, other researchers have also proposed different architectures with excellent performance. Zhang et al. proposed Variational NMT method, which has an innovative perspective in modeling translation task, and the corresponding experiment has indicated a better performance than the baseline of original NMT in Chinese-English and English-German tasks\cite{Zhang2016}. Zhou et al. have designed Fast-Forward Connections for RNN (LSTM), which can allow a deeper network in implementation and thus gets a better performance\cite{Zhou2016}. Shazeer et al. incorporated Mixture-of-Expert (MoE) architecture into the GNMT model, which has outperformed the original GNMT model\cite{Shazeer2017}. Concretely, MoE is one layer in the NMT model, which contains many sparsely combined experts (which are feed-forward neural networks in this experiment) and is connected with the RNN layer by a gate function. This method requires more parameters in total for the NMT model, but still maintains the efficiency in training speed. Since more parameters often imply a better representation ability, it demonstrates huge potential in the future.

\subsubsection {CNN based NMT}
Related work in trying other DNN models have also been proposed. Perhaps the most noted one is the Convolutional Neural Network (CNN) based NMT. In fact, CNN based models have also undergone many variations in its concrete architecture. But for a long while, most of these models can't have competitive performance with RNN based model, especially when the Attention Mechanism has emerged. 

In the development of CNN based NMT models,  Kalchbrenner \& Blunsom once tried a CNN encoder with RNN Decoder \cite{Kalchbrenner2013}, and it's maybe the earliest NMT architecture applied with CNN.  Cho et al. tried a gated recursive CNN encoder with RNN decoder, but it has shown worse performance than RNN encoder\cite{Cho2014b}. A fully CNN based NMT was proposed by Kaiser \& Bengio later\cite{Kaiser2016}, which applied Extended Neural GPU\cite{Kaiser2015}. The best performance in the early period of CNN based NMT was achieved by Gehring et al., which was a CNN encoder NMT and got the similar translation performance with RNN based model at that time\cite{Gehring2016}. Concurrently, Kalchbrenner et al. also proposed ByteNet (a kind of CNN) based NMT, which achieved the state-of-the-art performance on character-level translation but failed at word-level translation\cite{Kalchbrenner2016}. In addition, Meng et al. and Tu et al. proposed a CNN based model separately, which provides additional alignment information for SMT \cite{Meng2015}\cite{Tu2015}. 

Compared with RNN based NMT, CNN based models have its advantage in training speed; this is due to the intrinsic structure of CNN which allows parallel computations for its different filters when handling the input data. And also, the model structure has made CNN based models easier to resolve the gradient vanishing problem. However, there are two fatal drawbacks that affect their translation quality. First, since the original CNN based model can only capture the word dependencies within the width of its filters, the long dependency of words can only be found in high-level convolution layers; this unnatural  character often causes a worse performance than the RNN based model. Second, since the original NMT model compresses a sentence into a fixed size of the vector, a large performance reduction would happen when the sentence becomes too long. This comes from the limited representation ability in fixed size of the vector. Similar phenomenon can also be found in early proposed RNN based models, which are later alleviated by Attention Mechanism.

Some advanced CNN based NMT models have also been proposed with corresponding solutions in addressing the above drawbacks. Kaiser et al. proposed the Depthwise separable convolutions based NMT. The SliceNet they created can get similar performance with Kaiser et al. (2016) \cite{Kaiser2017}.  Gehring et al. (2017) followed their previous work by proposing a CNN based NMT that is cooperated with Attention Mechanism. It even got a better result than RNN based model\cite{Gehring2017}, but this achievement was soon outperformed by Transformer \cite{Vaswani2017}. 

\subsection {Encoder-Decoder Structure}
As is known, Encoder-Decoder is the most original and classic structure of NMT; it was directly inspired by NLM and proposed by Kalchbrenner \& Blunsom\cite{Kalchbrenner2013} and Cho et al.\cite{Cho2014b}. Despite all kinds of refinements in details and small tips, it was wildly accepted by almost all modern NMT models. Based on the discussion above, since RNN based NMT has held the dominant position in NMT, and to avoid being overwhelmed in describing all kinds of small distinctions between models' structures, we specifically focus our discussion just on the vanilla RNN based NMT, thus can help to trace back the development process of NMT.

The original structure of Encoder-Decoder structure is conceptually simple. It contains two connected networks (the encoder and the decoder) in its architecture, each for a different part of the translation process. When the encoder network receives one source sentence, it reads the source sentence word by word and compresses the variable-length sequence into a fixed-length vector in each hidden state. This process is called encoding. Then given the final hidden state of the encoder (referred to as thought vector), the decoder does the reverse work by transforming the thought vector to the target sentence word by word. Because Encoder-Decoder structure addresses the translation task from source data directly to the target result, which means there's no visible result in the middle process, this is also called end-to-end translation. The principle of Encoder-Decoder structure of NMT can be seen as mapping the source sentence with the target sentence via an intermediate vector in semantic space. This intermediate vector actually can represent the same semantic meaning in both two languages. 

For specific details of this structure, besides the model selection in the network, RNN based NMT models also differ in three main terms: (a) the directionality; (b) the type of activation function; and (c) the depth of RNN layer\cite{Luong2017}. In the following, we give a detailed description.

\textbf{Depth:}
For the depth of RNN, as we discussed in Section~\ref{sec:Sec2}, single layer RNN usually performs poorly comparing with multi-layer RNN. In recent years, almost all the models with competitive performance are using a deep network, which has indicated a trend of using a deeper model to get the state-of-the-art result. For example,  Bahdanau et al.\cite{Bahdanau2014} used four layers RNN in their model. 

However, simply increasing more layers of RNN may not always be useful. In the proof proposed by Britz et al.\cite{Britz2017}, they found that using 4 layers RNN in the encoder for specific dataset would produce the best performance when there is no other auxiliary method in the whole model. Besides that, stacking RNN layers may make the network become too slow and difficult to train. One major challenges is the gradient exploding and vanishing problem\cite{Bengio1994}, which will cause the gradient be amplified or diminished when processing back propagation in deep layers. Besides the additional gate structure in refined RNN (like LSTM and GRU), other methods have also been applied to alleviate this phenomenon. For example, in Wu et al.'s work, the residual connections are provided between layer, which can improve the value of gradient flow in the backward pass, thus can speed up the convergence process\cite{Wu2016}. Another possible problem is that a deeper model often indicates larger model capacity, which may perform worse on comparatively less training data due to the over-fitting.

\textbf{Directionality:}
In respect of directionality, a simple unidirectional RNN has been chosen by some researchers. For example, Luong et al. have directly used unidirectional RNN to accept the input sentence\cite{Luong2015}. In comparison, bidirectional RNN is another common choice that can empower the translation quality. This is because the model performance is affected by whether it 'knows' well about the information in context word when predicting current word. A bidirectional RNN obviously could strengthen this ability. 

In practice,  both Bahdanau et al. and Wu et al. used bidirectional RNN on the bottom layer as an alternative to capture the context information\cite{Bahdanau2014}\cite{Wu2016}.  In this structure, the first layer reads the sentence ``left to right'', and the second layer reads the sentence in a reverse direction. Then they are concatenated and fed to the next layer. This method generally has a better performance in experiment, although the explanation is intuitive: Based on the discussion of LM in Section~\ref{sec:Sec2}, the emergence probability of a specific word is determined by all the other words in both the prior and the post positions. When applying unidirectional RNN, word dependency between the first word and the last word is hard to be captured by the thought vector, since the model has experienced too many states in all time steps. On the country, bidirectional RNN provides an additional layer of information with reverse direction of reading words, which could naturally reduce this relative length within steps. 

The most visible drawback of this method is that it's hard to be paralleled, considering the time-consuming in its realization, both Bahdanau et al. and Wu et al. choose to apply just one layer bidirectional RNN in the bottom layer of the encoder, and other layers are all unidirectional layers\cite{Wu2016}\cite{Bahdanau2014}. This choice makes a trade-off between the feature representation ability with model efficiency, due to it can still enable the model to be distributed on multi GPUs\cite{Wu2016}. The basic concept of bidirectional RNN could find in Fig.~\ref{fig_env0}.

\textbf{Activation Function Selection:}
In respect of activation function selection, there are three common choices:  vanilla RNN, Long Short Term Memory (LSTM) \cite{Hochreiter1997}, and Gated Recurrent Unit (GRU) \cite{Cho2014b}. Comparing with the vanilla RNN, both the last two have some robustness in addessing the gradient exploding and vanishing problem\cite{Wu2016b}\cite{Bengio1994}.  Another sequence processing task has also indicated better performance achieved by GRU and LSTM\cite{Chung2014}. Besides, some innovative neural units have been proposed. Wang et al. proposed linear associative units, which can alleviate the gradient diffusion phenomenon in non-linear recurrent activation\cite{Wang2017}. More recently, Zhang et al. have created addition-subtraction twin-gated recurrent network (ATR). This type of unit reduces the inefficiency in NMT training and inference by simplifying the weight matrices among units \cite{Zhang2018}. All in all, in NMT task, LSTM is the most common choice.
\begin{figure}[b]
\centerline{\includegraphics[width=18pc]{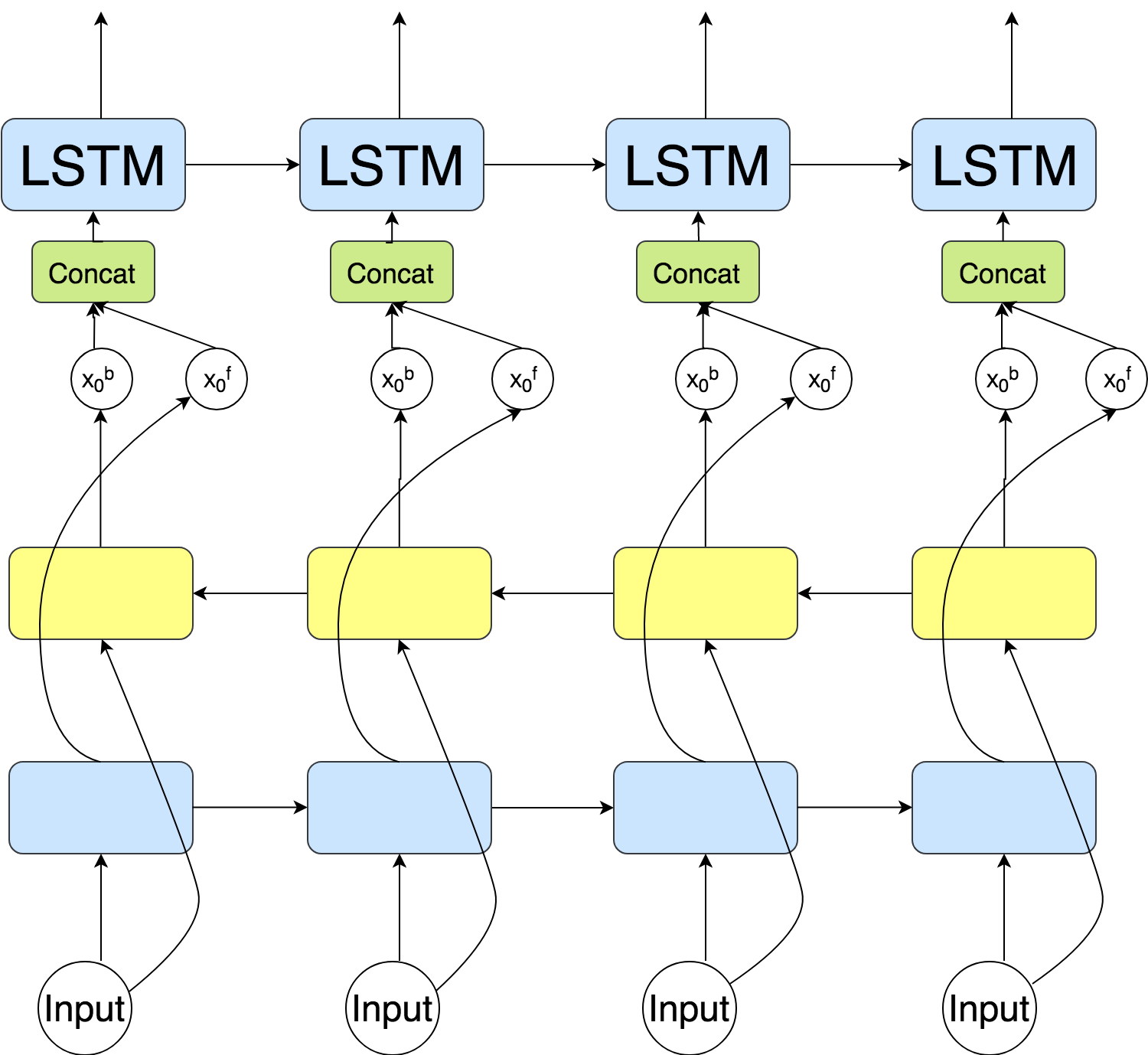}}
\caption{The concept of Bidirectional RNN}
%
\label{fig_env1}
\end{figure}

\subsection {Training method}
Before feeding the training data to the model, one pre-step is to transfer the words to vectors, which makes a proper form that the neural network could receive. Usually, the most frequent $V$ words in one language will be chosen, and each language generally has different word set. Despite that the embedding weights will be learned in the training period, the pre-trained word embedding vector such as word2vec\cite{Mikolov2013a}\cite{Mikolov2013b} or Glove vector\cite{Pennington2014} can also be applied directly.

In the training period, this model is fed by a bilingual corpus for Encoder and Decoder. The learning objective is to map the input sequence with the corresponding sequence in the target language correctly. Like other DNN models, the input sentence pair is embedded as a list of word vectors, and the model parameters are initialized randomly. The training process could be formulated as trying to updating its parameters periodically until getting the minimum loss of the neural network. In the implementation, RNN will refine the parameters after it processes a subset of data that contain a batch of training samples; this subset is called the mini-batch set. To simplify the discussion of the training process, we take one sentence pair (one training sample) as example. 

For the Encoder, the encoding RNN will receive one word in source sentence once a time-step. After several steps, all words will be compressed into the hidden state of the Encoder. Then the final vector will be transferred to the Decoder.

For Decoder, the input comes from two sources: the thought vector that is directly sent to Decoder, and the correct word in the last time-step (the first word is $<EOS>$). The output process in Decoder can be seen as a reverse work of Encoder; Decoder predicts one word in each time-step until the last symbol is $<EOS>$.

\subsection {Inference method}

After the training period, the model could be used for translation, which is called inference. The inference procedure is quite similar to the training process. Nevertheless, there is still a clear distinction between training and inference: at decoding time, we only have access to the source sentence, i.e., encoder hidden state. 

There is more than one way to perform decoding. Proposed decoding strategies include Sampling and  Greedy search, while the latter one is generally accepted and be evolved as Beam-search.

\subsubsection {General decoding work flow (greedy)}

The idea of greedy strategy is simple, as we illustrate in Fig.~\ref{fig_env3}. The Greedy strategy is only considering the predicted word with the highest probability. In the implementation of our illustration, the previously generated word would also be fed to the network together with the thought vector in the next time-step. The detailed steps are as follows:

\begin{figure}[t]
\centering
\includegraphics[width=18pc]{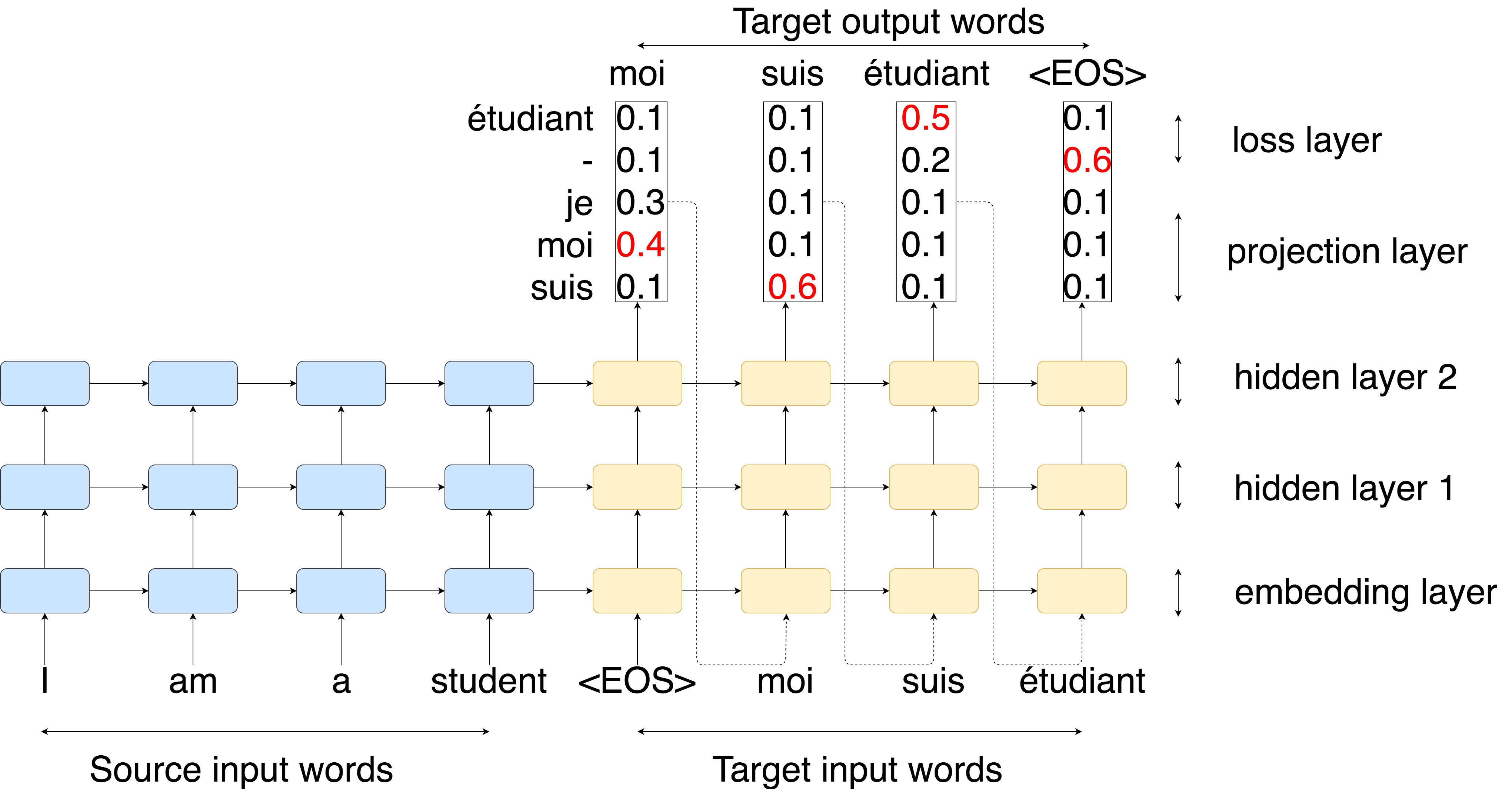}
\caption{The process of greedy decoding: each time the model would predict the word with highest probability, and use the current result as the input in next time step to get further prediction}
\label{fig_env3}
\end{figure}

1. The model still encodes the source sentence in the same way as during the training period to obtain the thought vector, and this thought vector is used to initialize the decoder.

2. The decoding (translation) process will start as soon as the decoder receives the end-of-sentence marker $<EOS>$ of source sentence.

3. For each time-step on the decoder side, we treat the RNN's output as a set of logits. We choose the word with the highest translation probability as the emitted word, whose ID is associated with the maximum logit value. For example, in Fig.~\ref{fig_env3}, the word ``moi'' has the highest probability in the first decoding step. We then feed this word as an input in the next time-step. The probability is thus conditioned on  the previous prediction (this is why we call it ``greedy'' behavior).

4. The process will continue until the ending symbol $<EOS>$ is generated as an output symbol.

\subsubsection {Beam-search}
While the Greedy search method has produced a pretty good result, Beam-search is a more elaborated one with better results. Although it is not a necessary component for NMT, Beam-search has been chosen by most of NMT models to get the best performance\cite{Britz2017}. 

The beam-search method was proposed by other sequence learning task with successful application\cite{Graves2012}\cite{Boulanger-Lewandowski2013}. It's also the conventional technique of MT task that has been used for years in finding the most appropriate translation result\cite{Och1999}\cite{Koehn2004}\cite{Chiang2005}. Beam-search can be simply described as retaining the top-$k$ possible translations as candidates at each time, where $k$ is called the beam-width. In the next time-step, each candidate word would be combined with a new word to form new possible translation. The new candidate translation would then compete with each other in log probability to get the new top-$k$ most reasonable results. The whole process continues until the end of translation.  

Concretely, the beam search can be formulated in the following steps:

\begin{algorithm}
\caption{Beam Search}
\label{alg:A}
\begin{algorithmic}
\STATE set Beamsize = K;
\STATE {$h_0 \Leftarrow encoder(S)$ } 
\STATE {$t \Leftarrow 1$ } 
\STATE // \quad $L_S$ means length of source sentence;
\STATE // \quad $\alpha$ is Length factor;
\WHILE{$n \leq \alpha*L_S$}
\STATE {$y_{1,i} \Leftarrow\ <EOS> $ } 
\WHILE{$i \leq K$} 
\STATE set $h_t \Leftarrow \ decoder(h_{t-1},y_{t,i})$;
\STATE set  $P_{t,i} = Softmax(y_{t,i})$;
\STATE set  $y_{t+1,i} \Leftarrow argTop \_ K(P_{t,i})$;
\STATE set $i=i+1$
\ENDWHILE
\STATE set $i=0$
\IF{$h_t == \ <EOS> $}
\STATE break;
\ENDIF
\STATE set $t=t+1$
\ENDWHILE
\STATE  select $argmax(p(Y))$ from $K$ candidates $Y_i$
\RETURN $Y_i$
\end{algorithmic}
\end{algorithm}

Besides the standard Beam-search which finds the candidate translation only by sorting log probability, this evaluation function mathematically tends to find shorter sentence. This is because a negative log-probability would be added at each decoding step, which lowers the scores with the increasing length of sentences\cite{Graves2013}. An efficient variant for alleviating this scenario is to add a length normalization\cite{Cho2014}. A refined length normalization was also proposed by Wu et al.\cite{Wu2016}.

Another kind of refined method in Beam-search is adding coverage penalty, which helps to encourage the decoder to cover the words in the source sentence as much as possible when generating an output sentence\cite{Wu2016}\cite{Tu2016}.

In addition, since this method finds $k$ times of translation(rather than one) until getting the final result, it generally makes the decoding process more time-consuming. In practice, an intuitive solution is to limit the beam-width as a small constant, which is a trade-off between the decoding efficiency and the translation accuracy. As reported by a comparison work, an experimental beam width for best performance is 5 to 10\cite{Britz2017}. 


\section{NMT with Attention Mechanism}
\label{sec:Sec4}
\subsection{Motivation of Attention Mechanism}
While the promising performance of NMT has indicated its great potential in capturing the dependencies inside the sequence, in practice, NMT still suffers a huge performance reduction when the source sentence becomes too long. Comparing with other feature extractors, the major weakness of the original NMT Encoder is that it has to compress one sentence into a fixed-length vector. When the input sentence becomes longer, the performance deteriorates because the final output of the network is a fixed-length vector, which may have limitation in representing the whole sentence and cause some information loss. And because of the limited length of vector, this information loss usually covers the long-range dependencies of words. While increasing the dimension of encoding vector is an intuitive solution, since the RNN training speed is naturally slow, a larger vector size would cause an even worse situation.

Attention Mechanism emerged under this circumstance. Bahdanau et al. \cite{Bahdanau2014} initially used this method as a supplement that can provide additional word alignment information in the decoding process, thus can alleviate the information reduction when the input sentence is too long. Concretely, Attention Mechanism is an intermediate component between Encoder and Decoder, which can help to determine the word correlation (word alignment information) dynamically. In the encoding period, it extends the vector of the final state in the original NMT model with a weighted average of hidden state in each time state, and a score function is provided to get the weight we mention above by calculating the correlation of each word in source sentence with the current predicting word. Thus the decoder could adapt its concentration in different translation steps by ordering the importance of each word correlation in source sentence, and this method can help to capture the long-range dependencies for each word respectively.

The inspiration for applying the Attention Mechanism on NMT comes from human behavior in reading and translating the text data. People generally read text repeatedly for mining the dependency within the sentence, which means each word has different dependency weight with each other. Comparing with other models in capturing word dependency information such as pooling layer in CNN or N-gram language model, attention mechanism has a global scope. When finding the dependency in one sequence, $N$-gram model will fix its the searching scope in a small range, usually the $N$ is equal to 2 or 3 in practice. Attention Mechanism, on the other hand, calculates the dependency between the current generating word with other words in source sentence. This more flexible method obviously bring a better result.

The practical application of Attention Mechanism is actually far beyond the NMT field, and it is even not an invention in NMT development. Some other tasks have also proposed similar methods that give weighted concentration on different position of input data, for example, Xu et al\cite{Xu2015}. proposed similar mechanism in handling image caption task, which can helps to dynamically locate different entries in image feature vector when generating description of them. Due to the scope of this survey, the following discussion would only focus on the Attention Mechanism in NMT.

\subsection{Structure of Attention Mechanism}
There are many variants in the implementation of Attention Mechanism. Here we just give the detailed description of Attention Mechanism which has been widely accepted as bringing significant contribution in the development of NMT.
\subsubsection{basic structure}
The structure of attention mechanism was originally proposed by Bahdanau et al. In later, Luong et al. proposed similar structure with small distinctions and extends this work\cite{Luong2015}\cite{Bahdanau2014}.

To simplify the discussion, here we take Luong et al.'s method as an example. Concretely, in encoding period, this mechanism receiving the input words like the basic NMT model, but instead of compressing all the information in one vector, every unit in the top layer of encoder will generate one vector that represents one time-step in the source sentence.

\begin{figure}[h]
\centering
\includegraphics[width=18pc]{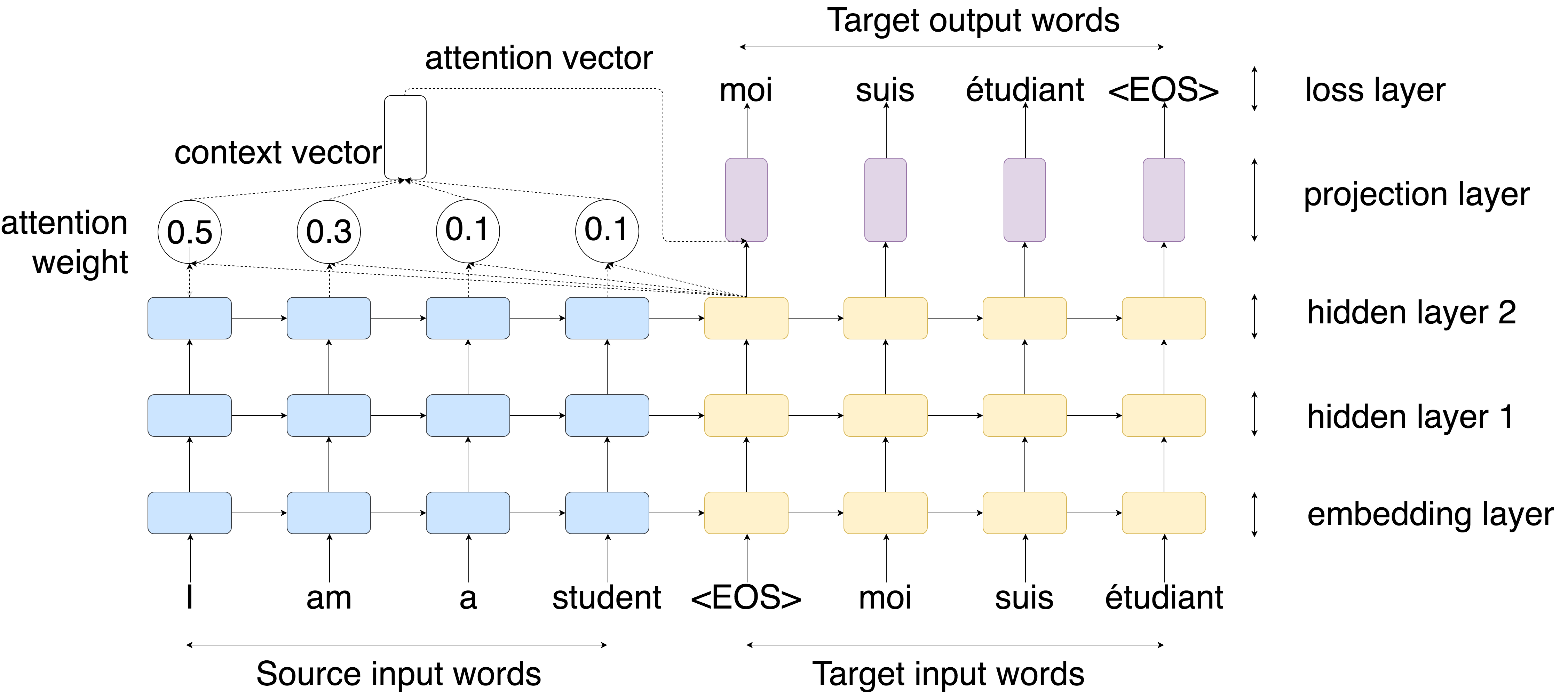}
\caption{The concept of Attention Mechanism,which can provide additional alignment information rather than just using information in fixed-length of vector}
%
\label{fig_env4}
\end{figure}

In the decoding period, the decoder won’t predict the word just use its own information. However, it collaborates with the attention layer to get the translation. 
The input of attention mechanism is the hidden states in the top layer of the encoder and the current decoder. It gets the relativity order by calculating the following steps:

1. The current decoding hidden state $h_t$ will be used to compare with all source states $h_s$ to derive the attention weights score $s_t$.

2. The attention weights $a_t$ is driven by normalization operation for all attention weight score.

3. Based on the attention weights, we then compute the weighted average of the source states as a context vector $c_t$.

4. Concatenate the context vector with the current decoding hidden state to yield the final attention vector(the exact combination method can be different).

5. The attention vector is fed as an input to decoder in the next time-step (applicable for input feeding). 
The first three steps can be summarized by the equations below:
\begin{equation}
\centering
 s_t = score(h_t,h_s)
\;\;\;\left[{Attention\  function}\right]
\label{eq5}
\end{equation}%
\begin{equation}
\centering
 a_t = \frac{exp (s_t)}{\sum\limits_{s=0}^{S} exp (s_t)}
\qquad\left[{Attention\ weight}\right]
\label{eq3}
\end{equation}%
\begin{equation}
\centering
 c_t = \sum\limits_{s}a_th_s   
 \;\;\qquad\qquad\qquad\left[{Context\  vector}\right] 
\end{equation}%

Among the above function, The score function could be defined in different ways. Here, we two classic definitions:
\begin{equation}
\centering
score(h_t,h_s) = \begin{cases}
 h_t^TWh_s  &[Luong's\  version]\\
 v_a^T tanh(W_1h_t,h_s) &[Bahdanau's\ version]
 \end{cases} 
\end{equation}%

Back to the decoding period, it receives the information from both two sides, the decoder hidden state and the attention vector, given the current two vectors, it then predicts the words by alignment them to a new vector, then it usually has another layer to predict the current target word.

\subsubsection{Global Attention \& Local Attention}

\textbf{Global Attention}

\begin{figure}[b]
\centering
\includegraphics[width=18pc]{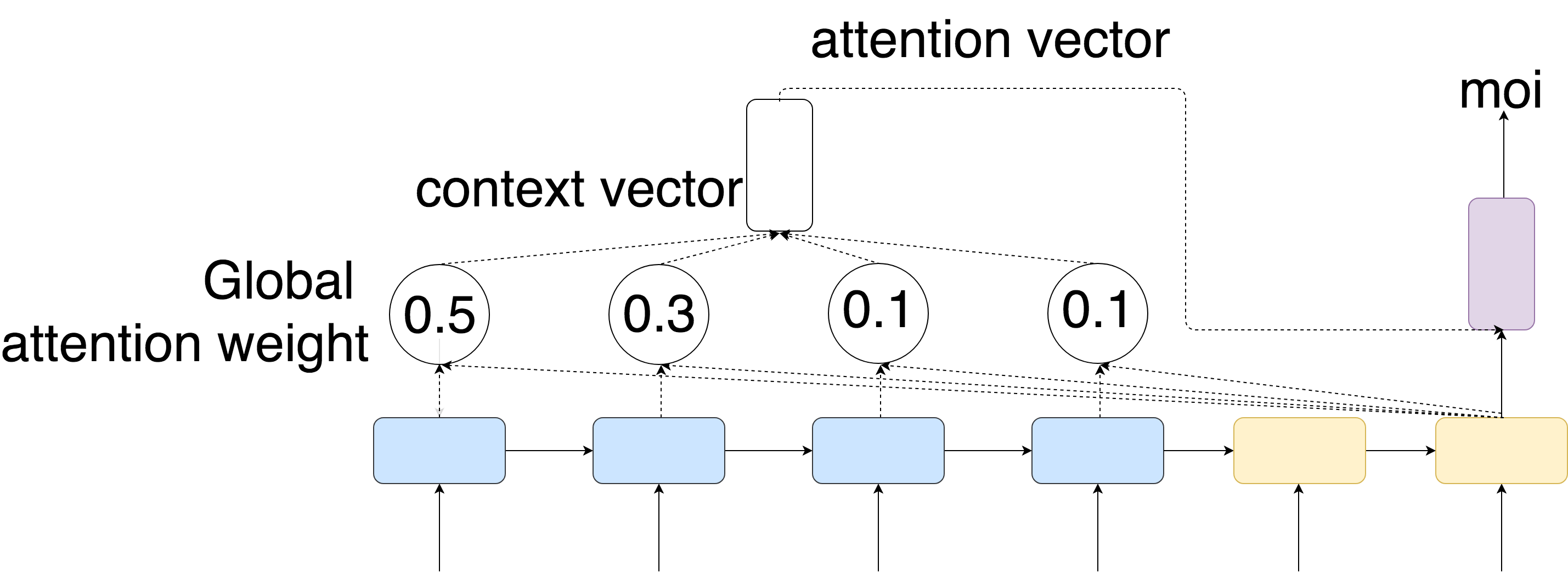}
\caption{The concept of Global attention, current decoder hidden state calculated with all the hidden states in source side to get the alignment information.}
%
\label{fig_env5}
\end{figure}

Global Attention is the method of Attention Mechanism we mentioned above, and it’s also a fluent type in various of Attention mechanism. The idea of Global Attention is also the original form of attention mechanism, though it got this name by Luong et al.\cite{Luong2015}, the corresponding term is Local Attention. The term —— ”global” derives from it calculates the context vector by considering the relevance order of all words in the source sentence. This method has excellent performance because more alignment information will generally produce a better result. A straightforward presentation in Fig.~\ref{fig_env5}

As we have introduced in Section~\ref{sec:Sec4}, this method considers all the source word in the decoding period. The main drawback is calculation speed deteriorates when the sequence is very long since one hidden state will be generated in one time-step in the Encoder, the cost of score function would be linear with the number of time-steps on the Encoder. When the input is a long sequence like a compound sentence or a paragraph, it may affect the decoding speed. 

\begin{figure}[h]
\centering
\includegraphics[width=18pc]{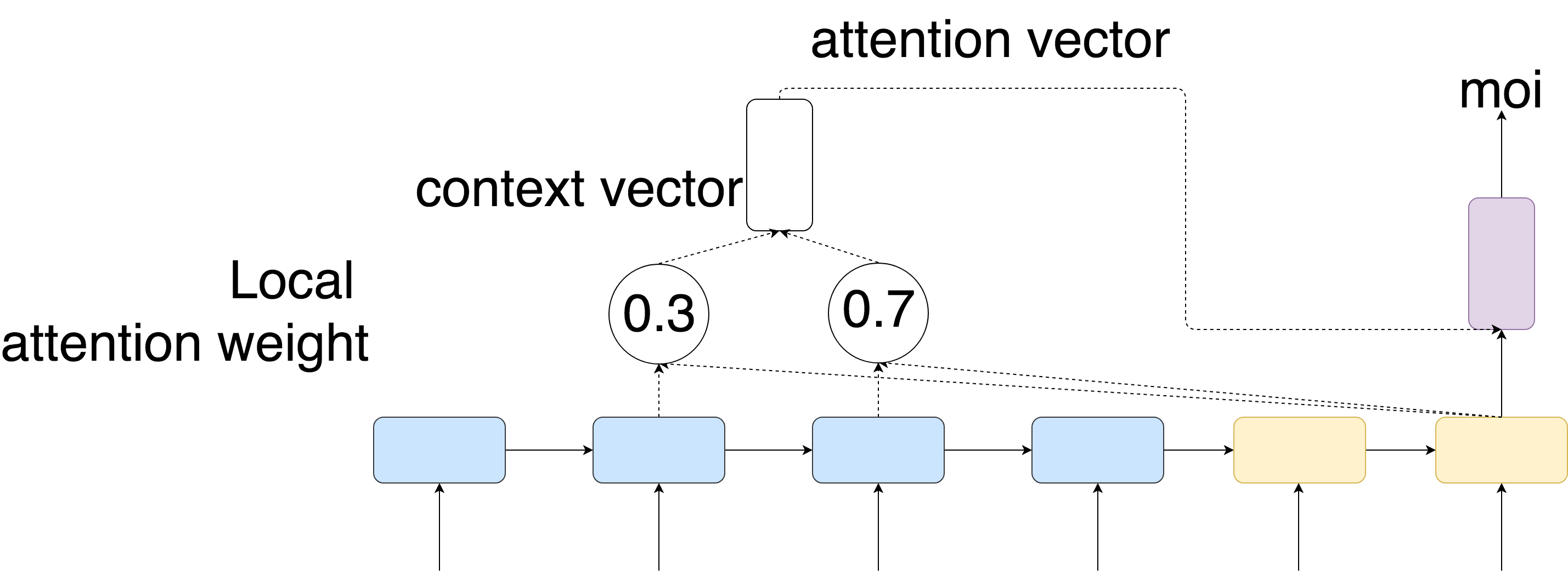}
\caption{The concept of Local attention, current hidden state calculated with a subset of all the hidden states in source side.}
%
\label{fig_env6}
\end{figure}

\textbf{Local attention}
Local attention was first proposed by Luong et al.\cite{Luong2015}. As illustrated in Fig.~\ref{fig_env6}, this model, on the other hand, will just calculate the relevance with a subset of the source sentence. Comparing with Global attention, it fixes the length of attention vector by giving a scope number, thus avoiding the expensive computation in getting context vectors.  The experiment result indicated that local attention can keep a balance between model performance with computing speed. 

The inspiration of local attention comes from the soft attention and hard attention in image caption generation task, which was proposed by Xu et al.\cite{Xu2015}. While the global attention is very similar to soft attention, the local attention, on the other hand, can be seen as a mixture method of soft attention with hard attention.

In theory, although covering more information would generally get a better result, the fantastic result of this method has indicated a comparable performance with global attention when it has been fine-tuned. This seems due to the common phenomenon in human language —— the current word would naturally have a high dependency with some of its nearby words, which is quite similar to the assumption of the n-gram language model.

In the details of calculation process, given the current target word’s position $p_t$, the model fixes the context vector in scope D. The context vector ct is then derived as a weighted average over the set of source hidden states within the range $[p_t-D,p_t+D]$; Scope D is selected by experience, and then it could be same steps in deriving the attention vector like Global attention.

\subsubsection{Input feeding approach}
Input feeding is a small tip in constructing the NMT structure, but from the perspective of providing alignment information, it can also be seen as a kind of attention.

The concept of input feeding is simple. In the decoding period, besides using the previously predicted words as input, it also uses the attention vector that in the previous time-step as additional input in next time-step\cite{Klein2017}\cite{Luong2015}. This attention vectors will concatenate with input vector to get the final input vector; then this new vector will be fed as the input in the next step.

\subsubsection{Attention Mechanism in GNMT }

GNMT is short for Google Neural Machine Translation, which is a well-known version of NMT with Attention Mechanism. GNMT was proposed by Wu et al.,\cite{Wu2016} and famous for its successful application in industrial level NMT system. With the help of many kinds of advanced tips in model detail, it got state-of-the-art performance at that time. Besides, the elaborate architecture of GNMT makes it have a better inference speed, which helps it more applicable in satisfying the industry need.

The concept of GNMT get the help of the current research in attention mechanism; it used Global Attention but was reconstruct by a more effective structure for model parallelization. The concrete details illustrated in the figure, it has two main points in this architecture. 
\begin{figure}[h]
\centering
\includegraphics[width=18pc]{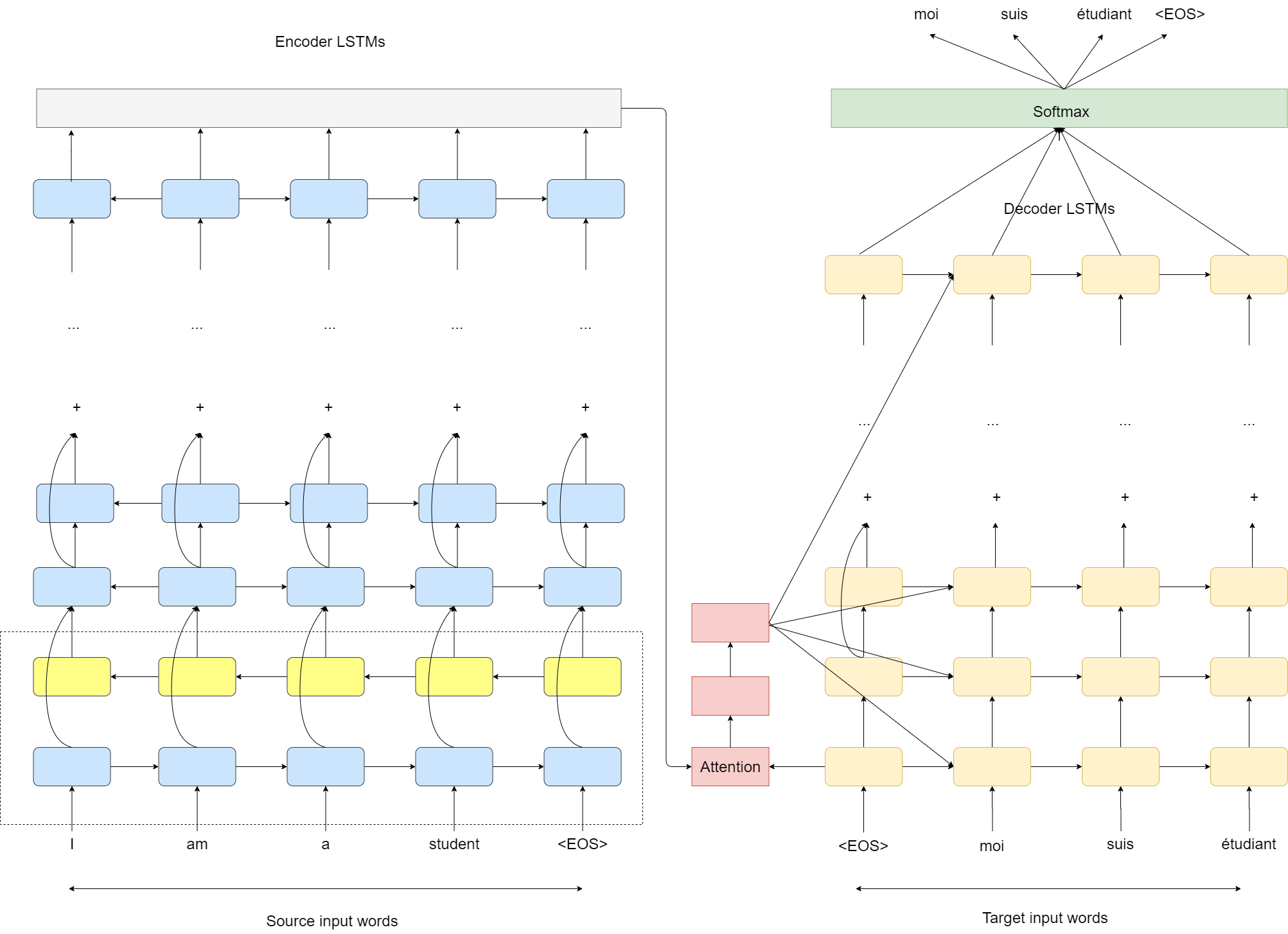}
\caption{Attention in GNMT, the Attention weight was driven by the bottom layer of Decoder and sent to all Decoder layers, which helps to improve computing parallelization}
%
\label{fig_env10}
\end{figure}
First, this structure has canceled the connection between the encoder and the decoder. So that it can have more freedom in choosing the structure of the encoder and decoder, for example, the encoder could choose the different dimensions in each layer regardless of the dimension in the decoder, only the top layer of the both encoder and decoder  should have same dimensions to guarantee that they can be calculated in mathematics for driving attention vector.

Second, this structure makes it easier for paralleling the model. Only the bottom layer of the decoder is used to get the context vector, then all of the remain decoding layers will use this context vector directly. This architecture can retain as much parallelism as possible.

For details of attention calculation, GNMT applying the Attention Mechanism like the way of calculating global attention, while the $score()$ function is a feed forward network with 1 hidden layer.

\subsubsection{Self-attention}
\begin{figure}[h]
\centering
\includegraphics[width=18pc]{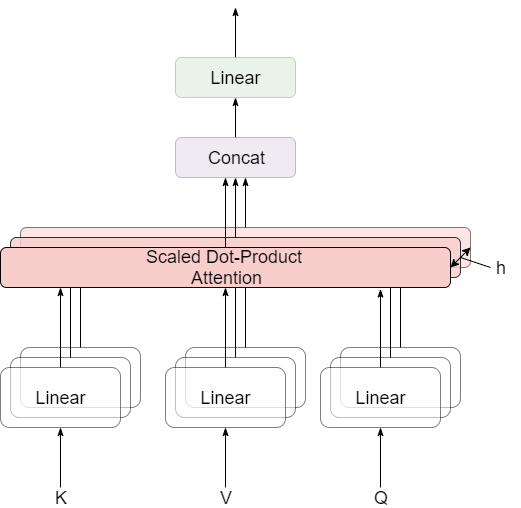}
\caption{The concept of Multi-head Self-attention}
%
\label{fig_env10}
\end{figure}
Self-attention is also called intra-attention, it is wildly known for its application in NMT task due to the emergence of Transformer. While other commonly noted Attention Mechanism driven the context information by calculating words dependency between source sequence with target sequence, Self-attention calculates the words dependency inside the sequence, and thus get an attention based sequence representation.

As for calculation steps, Self-attention first gets 3 vectors based original embedding for different purpose, the 3 vectors are  Query vector,  Key vector, and Value vector. Then the attention weights was calculated in this way:
\begin{equation}
\text { Attention }(Q, K, V)=\operatorname{softmax}\left(\frac{Q K^{T}}{\sqrt{d_{k}}}\right) V
\end{equation}
where the $\frac{1}{\sqrt{d_{k}}}$ is a scaled factor for avoiding to have more stable gradients that caused by dot products operation. In addition, the above calculation can be implemented in metrics  multiplication, so the words dependency can easily got in form of relation metrics. 

\subsection{Other related work}
Besides the above description in significant progress, there are also some other refinements in a different perspective of attention mechanism. 

In perspective of attention structure, Yang et al. improved the traditional attention structure by providing a network to model the relationship of word with its previous and subsequent attention\cite{Yang2016}. Feng et al. proposed a recurrent attention mechanism to improve the alignment accuracy, and it has been proved to outperformed vanilla models in large-scale Chinese–English task\cite{Feng2016}.

Moreover, other researches focus on the training process. Cohn et al. extended the original attention structure by adding several structural biases, they including positional bias, Markov conditioning, fertility, and Bilingual Symmetry\cite{Cohn2016},  model that integrated with these refinements have got better translation performance over the basic attention-based model. 
More concretely, the above methods can be seen as a kind of inheritance from the alignment model in SMT research, with more experiential assumption and intuition in linguistics:
$Position \ Bias:$ It assumed words in source and target sentence with the same meaning would also have a similar relative position, especially when both two-sentence have a similar word order.  As an adjustment of the original attention mechanism, it helps to improve the alignment accuracy by encouraging words in similar relative position to be aligned. Figure11111 demonstrated the phenomena strongly, where words in diagonal are tended to be aligned.
$Markov \  Condition:$ Empirically, in one sentence, one word has higher 
Correlation with its nearby words rather than those far from it, this is also the basement in explaining $context \ capture$ of n-gram LM. As for translation task, it's obvious that words are adjacent in source sentence would also map to the nearby position in target sentence, taking advantage of this property, this consideration thus improves the alignment accuracy by discouraging the huge jumps in finding the corresponding alignment of nearby words. In addition, the method with similar consideration but different implementation is local attention.
$Fertility:$ Fertility measures whether the word has been attended at the right level, it considers preventing both scenarios when the word hasn't got enough attention or has been paid too much attention. This design comes from the fact that the poor translation result is commonly due to repeatedly translated some word or lack coverage of other words, which refers to Under-translation and Over-translation.
$Bilingual \ Symmetry:$ In theory, word alignment should be a reversible result, which means the same word alignment should be got when translation processing form A  to  B with translation from B  to  A. This motivates the parallel training for the model in both directions and encouraging the similar alignment result.

The refinement infertility was further extended by Tu et al.\cite{Tu2016}, who proposed fertility prediction as a normalizer before decoding, this method adjusts the context vector in original NMT model by adding coverage information when calculating attention weights, thus can provide complementary information about the probability of source words have been translated in prior steps.  

Besides the intuition that heuristics from SMT, Cheng et al. applied the agreement-based learning method on NMT task, which encourages joint training in the agreement of word alignment with both translation directions \cite{Cheng2015}. In later, Mi et al. proposed a supervised method for attention component, and it utilized annotated data with additional alignment constraints in its objective function, experiments in Chinese-to-English task has proven to benefit for both translation performance and alignment accuracy\cite{Mi2016a}.
\section{Vocabulary Coverage Mechanism}
\label{sec:Sec5}
Besides the long dependency problem in general MT tasks, the existence of unknown words is another problem that can severely affect the translation quality. Different from traditional SMT methods which support enormous vocabulary, most of NMT models suffer from the vocabulary coverage problem due to the nature that it can only choose candidate words in predefined vocabulary with a modest size. In terms of vocabulary building, the chosen words are usually frequent words, while the remaining words are called unknown words or out-of-vocabulary (OOV) words.

Empirically speaking, the vocabulary size in NMT varies between 30k-80k at most in each language, with one marked exception was proposed by Jean et al., who once used an efficient approximation for $softmax$ to accommodate for the immense size of vocabulary (500k)\cite{Jean2015}. However, the vocabulary coverage problem still persists widely because of the far more number of OOV words in $de\ facto$ translation task, such as proper nouns in different domains and a great number of rarely used verbs. 

Since the vocabulary coverage in NMT is extremely limited, handling the OOV words is another research hot spot. This section demonstrates the intrinsic interpretation of the vocabulary coverage problem in NMT and the corresponding solutions proposed in the past several years. 

\subsection{Description of Vocabulary Coverage problem in NMT}
\begin{figure}[h]
\centering
\includegraphics[width=18pc]{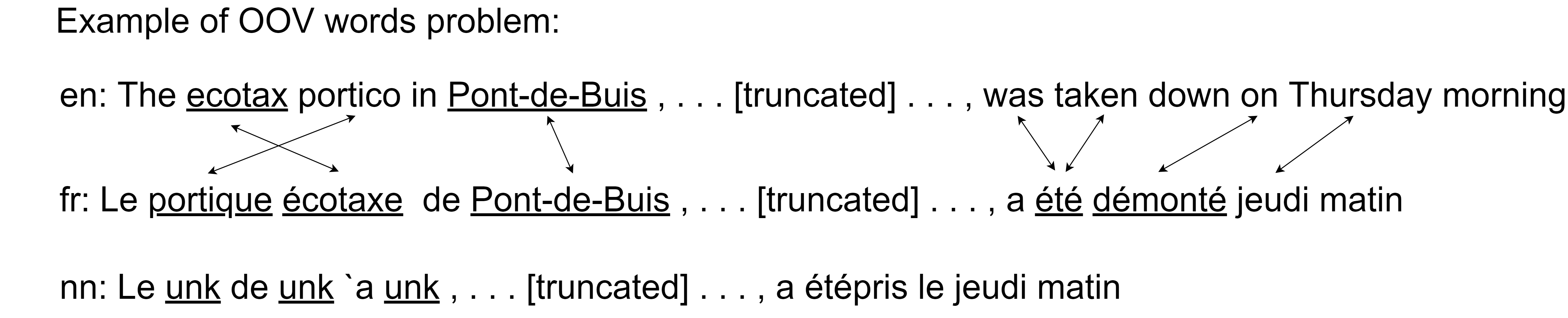}
\caption{An example of OOV words problem presented in\cite{Luong2015}. $en$ and $fr$ denote the source sentence in English and the  corresponding target sentence in French, $nn$ denotes the neural network's result.}
%
\label{fig_env11}
\end{figure}

Based on the scenario as mentioned above, in the practical implementation of NMT, the initial way is choosing a small set of vocabulary and converting a large number of OOV words to one uniform ``UNK'' symbol (or other tags) as illustrated in Fig.~\ref{fig_env11}.  This intuitive solution may hurt translation performance in the following two aspects. First, the existence of ``UNK'' symbol in translation may hurt the semantic completeness of sentence; ambiguity may emerge when ``UNK'' replace some crucial words\cite{Gulcehre2016}. Second, as the NMT model hard to learn information from OOV words, the prediction quality beyond the OOV words may also be affected\cite{Jiajun2016}.

\begin{table}[htbp]
  \centering
  \caption{BLEU Performance of NMT Models}
    \begin{tabular}{l|c|c}
    \toprule
    \multicolumn{1}{c|}{\multirow{2}[4]{*}{Model}} & \multicolumn{2}{c}{BLEU} \\
\cmidrule{2-3}          & EN-DE & EN-FR \\
    \midrule
    ByteNet & 23.75 &  \\
    Deep-Att + PosUnk &       & 39.2 \\
    GNMT + RL & 24.6  & 39.92 \\
    ConvS2S & 25.16 & 40.46 \\
    MoE   & 26.03 & 40.56 \\
    \midrule
    Deep-Att + PosUnk Ensemble &       & 40.4 \\
    GNMT + RL Ensemble & 26.3  & 41.16 \\
    ConvS2S Ensemble & 26.36 & 41.29 \\
    \midrule
    Transformer (base model) & 27.3  & 38.1 \\
    Transformer (big)  & 28.4  & 41.8 \\
    \bottomrule
    \end{tabular}%
  \label{tab:addlabel}%
\end{table}%

Besides the unsurprising observation that NMT performed poorly on sentence with more OOV words than with more frequent words, some other phenomena in MT task are also hard to be handled like multi-word alignment, transliteration, and spelling, .etc. \cite{Sutskever2014}\cite{Bahdanau2014}. They are seen as a similar phenomenon which is also caused by unknown words problem or suffers from rare training data\cite{Sennrich2015}.

For most of NMT model, choosing a modest size of vocabulary list is virtually a trade-off between the computation cost with translation quality. Also, it has been found the same thing in training NLM\cite{Morin2005}\cite{Mnih2009}. Concretely, the computation cost mainly comes from the nature of the method in getting predicting word — a normalization operation, which is used repeatedly in training of the DL model. Specifically, in NMT task, since DL model needs to adjust the parameters each time, the probability of current word thus would be calculated repeatedly to get the gradient, and since NMT model calculates the probability of current word when making a prediction, it needs to normalize the all of words in the vocabulary each time. Unfortunately, the normalization process is time-consuming due to its time complexity is linear with the vocabulary size, and this attribute has rendered the same time complexity in the training process.

\subsection{Different Solutions}
Related researches have proposed various methods in both the training and inference process. Further, these methods can be roughly divided into three categories based on their different orientations. The first one is intuitively focused on finding solutions in improving computation speedup, which could support a more extensive vocabulary. The second one focus on using context information, this kind of method can address some of the unknown words(such as Proper Noun) by copying them to translation result as well as low-frequency words which cause a poor translation quality. The last one, which is more advanced, prefers to utilize information inside the word such as characters, because of their flexibility in handling morphological variants of words, this method can support translating OOV words in a more ''intelligent" way.

\subsubsection{methods by computation speedup}
For computation speedup method, there are lots of literature that implement their idea in NLM training. The first thought in trying computation speedup is to scale the $softmax$ operation. Since an effective $softmax$ calculation could obviously support a larger vocabulary, this kind of trying has got a lot of attention in NLM literature. Morin \& Bengio \cite {Morin2005} proposed hierarchical models to get an exponential speedup in the computation of normalization factor, thus help to accelerate the gradient calculation of word probabilities. In concrete details, the original model has transformed vocabulary into a binary tree structure, which was built with pre-knowledge from WordNet\cite{Miller1998}. 

The initial experiment result shows that this hierarchical method is comparable with traditional trigram LM but fails to exceed original NLM; this is partly because of utilizing handcrafted feature from WordNet in the tree building process.  As a binary tree can provide a significant improvement in cost-effective between the speed with performance, further work still focuses on this trend to find better refinement.  Later on, Mnih \& Hinton followed this work by removing the requirement of expert knowledge in tree building process\cite{Mnih2009}. 

A more elegant method is to retain the original model but change the method in calculating the normalization factor. Bengio \& Sen´ecal proposed importance sampling method to approximate the normalization factor\cite{BengioSenécal2003}.  However, this method is not stable unless with a careful control\cite{BengioSenécal2008}. Mnih \& Teh used noise-contrastive estimation to learn the normalization factor directly, which can be more stable in the training process of NLM  \cite{Mnih2012}. Later, Vaswani et al. proposed a similar method with application in MT\cite{Vaswani2013}.

The above methods are difficult to be implemented parallelly by GPUs. Further consideration found solutions that are more GPU friendly. Jean et al. alleviated the computation time by utilizing a subset of vocabulary as candidate words list in the training process while used the whole vocabulary in the inference process. Based on the inspiration of using importance sampling in earlier work\cite{BengioSenécal2008},  they proposed a pure data segmentation method in the training process. Specifically, they pre-processed the training data sequentially, choosing a subset of vocabulary for each training example with the number of its distinct words reached threshold $t$(which is still far less than the size of original vocabulary). In the inference process, they still abandon using the whole vocabulary and proposing a hybrid candidate list alternatively.  They composed candidate words list from two parts. The first part is some specific candidate target words that translated from a pre-defined dictionary with the others are the K-most frequent words. In the practical performance analysis, this method remains the similar modest size of candidate words in the training process; thus, it can maintain the computational efficiency while supporting an extremely larger size of candidate words\cite{Jean2015}. Similarly, Mi et al. proposed vocabulary manipulation method which provides a separate vocabulary for different sentences or batches, it contains candidate words from both word-to-word dictionary and phrase-to-phrase library\cite{Mi2016b}.

Besides all kinds of corresponding drawbacks in the above method, the common weakness of all these methods is they still suffer from the OOV words despite a larger vocabulary size they can support. This is because the enlarged vocabulary is still size limited, and there's no solution for complementary when encountering unknown words, whereas the following category of methods can partly handle it. In addition, simply increasing the vocabulary size can merely bring little improvement due to the Zipf’s Law, which means there is always a large tail of OOV words need to be addressed\cite{Gulcehre2016}.

\subsubsection{methods by using context information}
Besides the above variants which focus on computation speed-up, a more advanced category is using context information. Luong et al. proposed a word alignment algorithm which collaborates with Copy Mechanism to post-processing the translation result. This old but useful operation was inspired by the common word(phrase) replacement method in SMT and has achieved a pretty considerable improvement in BLEU \cite{Luong2014}. Concretely, in Luong's method, for each of the OOV words, there's a ``pointer'' which map to the corresponding word in the source sentence. In the post-processing stage, a predefined dictionary was provided with ``pointer'' to find the corresponding translation, while using directly copy mechanism to handle the OOV words that not in the dictionary. 

The popularity of Luong et al. 's method is partly because the Copy Mechanism actually provides an infinite vocabulary. Further research has refined this alignment algorithm for better replacement accuracy and generalization. Choi et al. extended Luong et al.'s approach by dividing OOV words into one of three subdivisions based on their linguistic features. \cite{Choi2017} This method can help to remap the OOV words effectively. Gulcehre et al. done several refinements in this category, they applied Copy Mechanism similar to Luong et al. but cooperate the Attention Mechanism in determining the location of word alignment, which is more flexible in addressing alignment and could be directly utilized in other tasks which alignment location varies dramatically in both sides (like text summarization). Besides that, they synthesized Copy Mechanism with general translation operation by adding a so-called switching network to decide which operation should be applied in each time-step, this could be thought to improve the generalization of the whole model. \cite{Gulcehre2016}. Gu et al. made parallel efforts in integrating different mechanisms, they proposed a kind of Attention Mechanism called CopyNet with the vanilla encoder-decoder model, which can be naturally extended to handle OOV words in NMT task\cite{Gu2016}. Additionally, they found that the attention mechanism has driven more by the semantics and language model when using traditional word translation, but by location when using copying operation.

\begin{figure}[h]
\centering
\includegraphics[width=18pc]{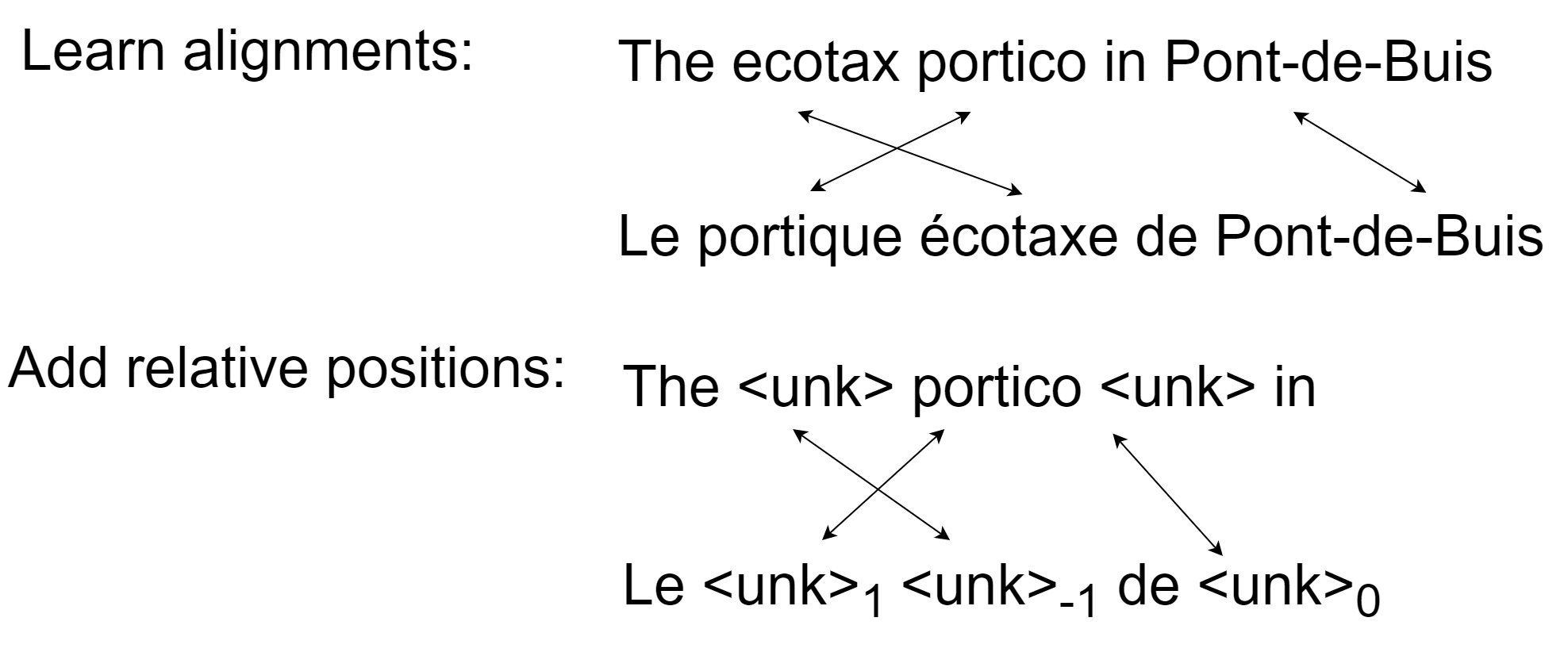}
\caption{An example of one kind of Copy Mechanism proposed by Luong et al.\cite{Luong2015}, the subscripts of $<unk>$ symbol (refer as $d$)is a relative position of corresponding source words, where the alignment  relation is a target word at position $j$ is aligned to a source word at position $i$ $=$ $j$ $+$ $d$.  }
%
\label{fig_env111}
\end{figure}

Besides the Copy Mechanism, using extra knowledge is also useful in handling some other linguistic scenarios, which is highly related to the vocabulary coverage problem. Arthur et al. incorporated lexicons knowledge to assist with translation in low-frequency words\cite{Arthur2016}. Feng et al. proposed a similar method with a memory-augmented NMT (M-NMT) architecture, and it used novel attention mechanism to get the extra knowledge from the dictionary that constructed by SMT\cite{Feng2017}. Additionally, using context information can also be applied to improve the translation quality of ambiguous words (Homographs)\cite{Liu2017}. 

In a nutshell, there are many context-based refinements have been proposed, most of them using Copy Mechanism to handle the OOV words with various of alignment algorithm to locate the corresponding word in the target side. However, these kinds of methods have a limited room for further improvement because the Copy Mechanism is too crude to handle the sophisticated scenarios in different languages. Practically, these methods perform poorly in languages which are rich morphology like Finnish and Turkish, which motivated method with better generalization\cite{Ling2015b}.

\subsubsection{Methods in fine grit level}

This sub-section introduce some more ``intelligent'' ways that focus on using additional information inside the word unit. It's obviously that such additional information could enhance the ability in covering the various of the linguistic phenomenon.

In previous research, although using semantic information of word unit could provide the vast majority of learning features, other features in sub-word level are generally ignored. From the perspective of linguistic, the concept of ``word'' is the basic unit of language but not the minimum one in containing semantic information, and there are abundant experienced rules could be learned from the inside of word units like shape and suffix. Comparing with identity copy or dictionary translation which regards the rare words as an identical entity, the refined method using fine-grit information is more adaptable. Further, a more ``radical'' method was proposed, which just treats words in character level completely. It would be an innovative concern for future work.

In this category, one popular choice is using the sub-word unit, the most remarkable achievement was proposed by Sennrich et al., which has been proved to have the best performance in some shared result\cite{Sennrich2015}. Concretely, in this design, they treat unknown words as sequences of sub-word units, which is reasonable in terms of the composition of the vast majority of these words(e.g., named entities, loanwords, and morphologically complex words). In order to represent these OOV words completely, one intuitive solution is to build a predefined sub-word dictionary that contains enough variants of units. However, restoring the sub-words cause massive space consumption in vocabulary size, which is effectively cancels out the whole purpose of reducing the computational efficiency in both time and space. Under this circumstances, a Byte Pair Encoding (BPE) based sub-words extraction method was applied for word segmentation operation in both sides of languages, which successfully adapted this old but effective data compression method in text pre-processing. 

In concrete details of this adapted BPE method, it alternated merging frequent pairs of bytes with characters or sequence of characters, and word segmentation process followed the below steps:

(1) Prepare a large training corpus(generally are bilingual corpus).

(2) Determined the size of the sub-word vocabulary.

(3) Split the words to a sequence of characters (using a special space character for marking the original spaces). 

(4) Merge the most frequent adjacent pair of characters (e.g., in English, this may be c and h into ch). 

(5) Repeat step 4 until reaching the fixed given number of times or the defined size of the vocabulary. Each of these steps would increase the vocabulary by one.

The figure shows a toy example of the method. As for the practical result of word segmentation, the most frequent words will be merged as single tokens, while the rare words(which is similar to the OOV words in previous categories' work) may still contain un-merged sub-words. However, they have been found rare in processed text\cite{Sennrich2015}.

Further work of BPE method has also been proposed to obtaining a better generalization. Taku proposed subword regularization in later as an alternative to handle the spurious ambiguity phenomenon in BPE, and they further proposed a new subword segmentation algorithm based on a uni-gram language model, which shares the same idea with BPE but was more flexible in getting multiple segmentation based on their probabilities.  Similarly, Wu et al. used ``workpieces'' concept to handle the OOV words which were once applied on the Google speech recognition system to solve Japanese/Korean segmentation problem\cite{Schuster2012}\cite{Wu2016}. This method breaks words into word pieces to get a balance between flexibility with efficiency when using single characters and full words separately. 

Another notable concern is modeling sequence in characters-level. Inspired by using character-level information completely in building NLM\cite{Kim2016}, Costa-jussa` \& Fonollosa used CNN with highway network to model the characters directly\cite{Costa-Jussa2016}, they deployed this architecture in source side with a common word-level generation in target side. Similarly, Ling et al. and Ballesteros et al. have proposed model respectively that using RNN(LSTM) to build character level embedding and composes it into the word embedding \cite{Ling2015b} \cite{Ballesteros2015}, this idea later has been applied in building an RNN based character-level NMT model \cite{Ling2015a}. More recently, Luong and Manning(2016) proposed a hybrid model that combines the word level RNN with character level RNN for assist\cite{Luong2016}. Concretely, Luongs' method translates mostly at the word level, when encounter an OOV word, character level RNN would be used for the consult. The figure shows the detailed architecture of this model.

On the other hand, the trying of designing a fully character-level translation model has also got attention accordingly. Chung et al. used BPE method to extract a sequence of subword in encoder side, they just varied the decoder by using pure characters, and it has indicated to provide comparable performance with models uses sub-words \cite{Chung2016}.  Motivated by aforementioned work, Lee et al. proposed fully character-level NMT without any segmentation, it was based on CNN pooling with highway layers, which can solve the prohibitively slow speed of training in Luong and Manning's work \cite{Lee2017}.

\section{Advanced models}
\label{sec:Sec6}
This section gives a demonstration of some advanced models that have got the state-of-the-art performance, while all of them belong to different categories of model structure. Experimental result indicated that all these networks can achieve similar performance with different advantages in their corresponding aspects.

\subsection{ConvS2S}
ConvS2S is short for Convolutional Sequence to Sequence, which is an end-to-end NMT model proposed by Gehring et al.\cite{Gehring2017}. Different with most of RNN based NMT models, ConvS2S is entirely CNN based model both in encoder and decoder. In the network structure, ConvS2S stacked 15 layers of CNN in its encoder and decoder with fixed kernel width of 3. This deep structure helps to mitigate the weakness in capturing context information.

In respect of network details, ConvS2S applied Gated Linear Units (GLU)\cite{Dauphin2017} in building network, which provide a gated function for output of convolution layer. Specifically, the output of convolution layer $Y \in \mathbb{R}^{2 d}$ which is a vector with double times dimensions (2$d$ numbers of dimensions) of each input element's embedding ($d$ numbers of dimensions), the gated function processes the output $Y=[A B] \in \mathbb{R}^{2 d}$ by implementing the equation $8$, where both $A$ and $B$ are $d$ dimensions vector, and the function $\sigma(B)$ is a gated function used to control which inputs $A$ of the current context are relevant. This non-linearity operation has been proved to be more effective in applying training language model\cite{Dauphin2017} , surpassing those only applying $tanh$ function on $A$ \cite{Oord2016}. In addition, ConvS2S also used residual connection \cite{He2015} between different convolution layers.
\begin{equation}
v([A B])=A \otimes \sigma(B)
\end{equation}
Besides the innovation of CNN based encoder-decoder structure, ConvS2S also applied similar Attention Mechanism that has been wildly accepted by RNN model, called Multi-step Attention. Concretely, Multi-step Attention is a separate attention structure applied in each decoder layer. In the process of calculating attention, the current hidden state $d_{i}^{l}$ (i.e., the output of the $l$th layer) has combined with previous output embedding $g_{i}$ as a vector of decoder state summary  $d_{i}^{l}$:
\begin{equation}
d_{i}^{l}=W_{d}^{l} h_{i}^{l}+b_{d}^{l}+g_{i}
\end{equation}
Then the attention vector $a_{i j}^{l}$ (i.e., the attention of state $i$ with source element $j$ in decoder layer $l$) would be driven by the dot-product of the summary vector with the output of the final encoder layer $z_{j}^{u}$. 
\begin{equation}
a_{i j}^{l}=\frac{\exp \left(d_{i}^{l} \cdot z_{j}^{u}\right)}{\sum_{t=1}^{m} \exp \left(d_{i}^{l} \cdot z_{t}^{u}\right)}
\end{equation}
Lastly, the context vector is calculated as the weighted average of the attention vector $a_{i j}^{l}$ with the encoder output $z_{j}^{u}$ as well as the encoder input $e_{j}$.
\begin{equation}
c_{i}^{l}=\sum_{j=1}^{m} a_{i j}^{l}\left(z_{j}^{u}+e_{j}\right)
\end{equation}
\begin{figure}[h]
\centering
\includegraphics[width=18pc]{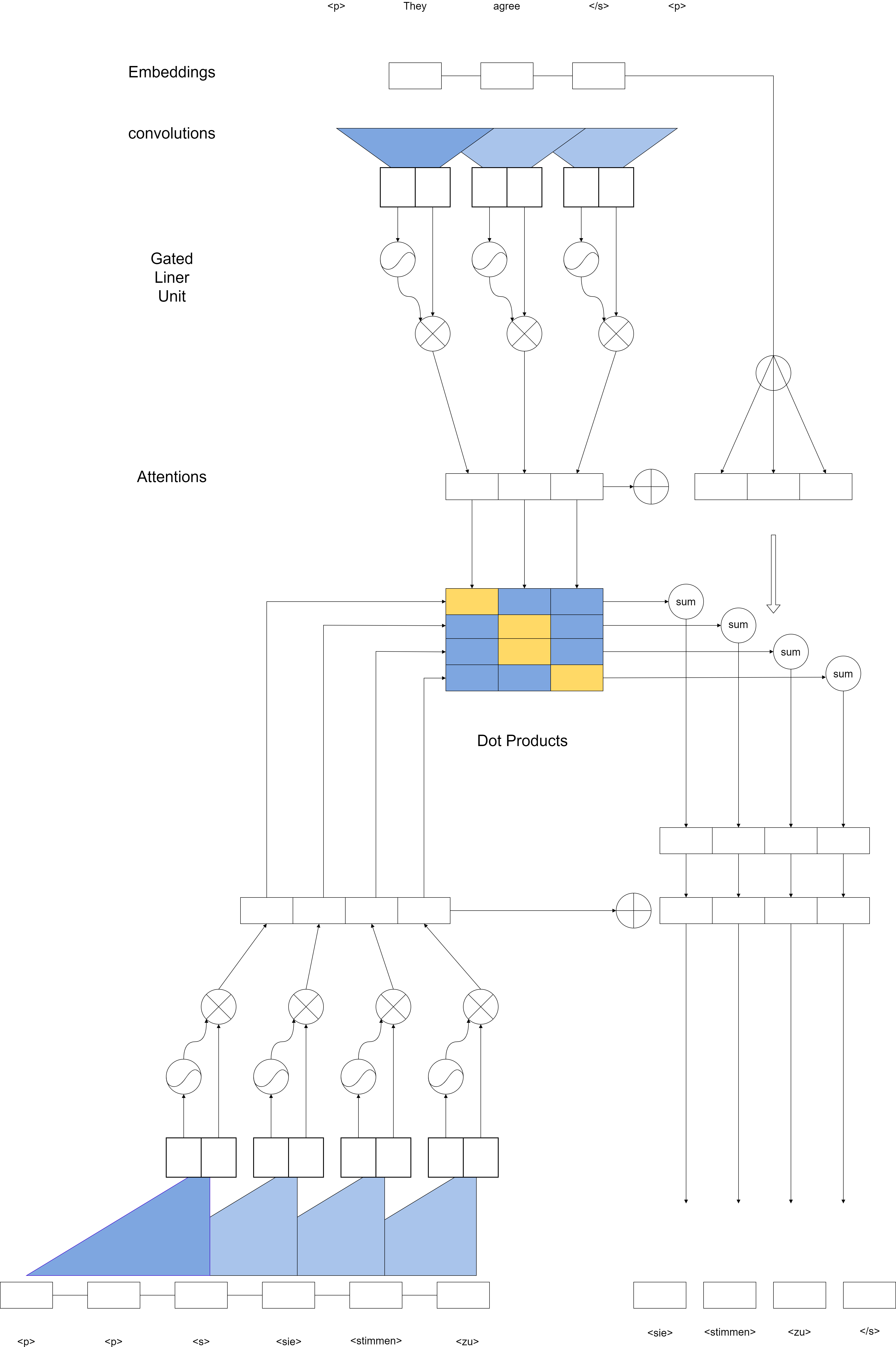}
\caption{The structure of ConvS2S model, a successful CNN based NMT model with competitive performance to the state-of-the-art }
%
\label{fig_env12}
\end{figure}
\subsection{RNMT+}
RNMT+ was proposed by Chen et al.\cite{Chen2018}. This model has directly inherited the structure of GNMT model that was proposed by Wu et al.\cite{Wu2016}. Specifically, RNMT+ can be seen as an enhanced GNMT model, which demonstrated the best performance of RNN based NMT model. In model structure, RNMT+ mainly differs from the GNMT model in the following several perspectives:

First, RNMT+ used six bi-directional RNN (LSTM) in its decoder, whereas GNMT used one layer of bi-directional RNN with seven layers of unidirectional RNN. This structure has sacrificed the computation efficiency in return for the extreme performance.

Second, RNMT+ applied Multi-head additive attention instead of the single-head attention in conventional NMT model, which can be seen as taking advantage of Transformer model.

Third, synchronous training strategy was provided in the training process, which improved the convergence speed with model performance based on empirical results\cite{Chen2016}. 

In addition, inspired by Transformer model, per-gate layer normalization \cite{Ba2016} was applied, which has indicated to be helpful in stabilizing model training. 
\begin{figure}[h]
\centering
\includegraphics[width=18pc]{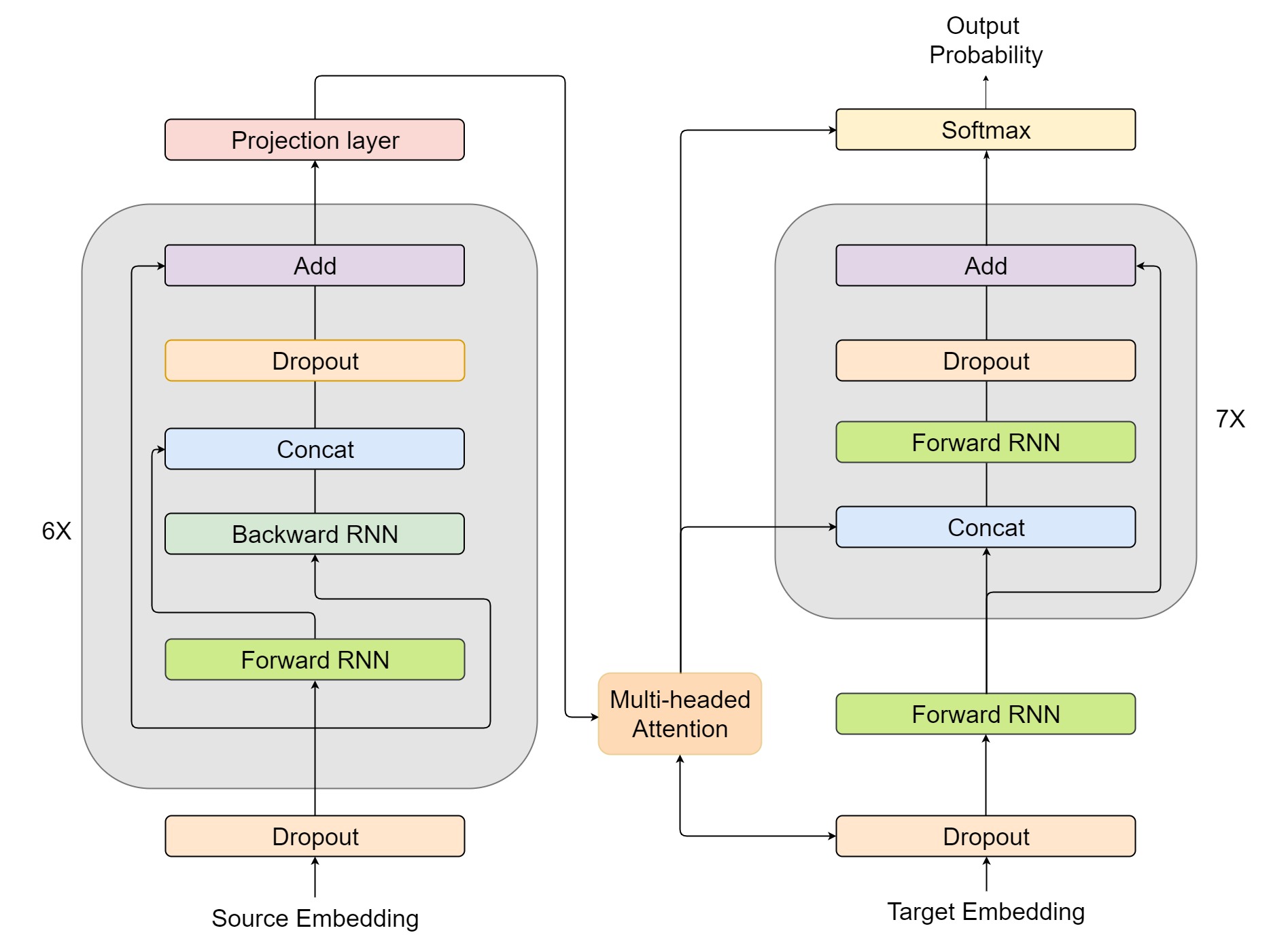}
\caption{The structure of RNMT+ model,which has the similar structure of GNMT with adaptive innovation in Attention Mechanism} 
%
\label{fig_env13}
\end{figure}
\subsection{Transformer and Transformer based models}
Transformer is a new NMT structure proposed by Vaswani et al.\cite{Vaswani2017}. Different from existing NMT models, it has abandoned the standard RNN/CNN structures and designed an innovative multi-layer self-attention blocks that are incorporated with a positional encoding method. This new trend of structure design takes the advantages from both RNN and CNN based model, which has been further used for initializing the input representation for other NLP tasks. Notably, Transformer is a complete Attention based NMT model.

\subsubsection{Structure of the model}
Transformer has its distinct structure in its model, where the major differences are the input representation and multi-head attention.

(1) Input representation

Transformer has its unique representation in handling input data that quite different with recurrent or convolution model. For computing Self-Attention we mentioned, transformer handles the input as three kind of vectors for different purpose. They are Key,Value and Query vectors. And all these vectors are driven by multiplying the input embedding with three matrices that we trained during the training process.

Also, there's Positional Encoding method was applied for enhancing the modeling ability of sequence order, since Transformer has abandoned recurrence structure, this kind of method made a compensation by inject word order information in to feature vector, which can avoid the model to become invariant to sequence ordering \cite{Parikh2016}. Specifically, Transformer add Positional Encoding to the input embedding at the bottoms of the encoder and decoder stacks, the Positional Encoding has been designed to have the same dimension with model embedding and thus could be summed. Positional Encoding can be calculated by applying positional functions directly or be learned \cite{Gehring2017}, with be proven to have similar performance in final evaluation. In Transformer, adding Positional Encoding by using $since$ and $cosine$ functions is finally chosen, and each position can be encoded in the following way:

\begin{equation}
\begin{array}{r}{P E_{(p o s, 2 i)}=\sin \left(p o s / 10000^{\left.2 i / d_{\text {model }}\right)}\right.} \\ {P E_{(\text {pos, } 2 i+1)}=\cos \left(\text {pos} / 10000^{2 i / d_{\text {model }}}\right)}\end{array}
\end{equation}
where $pos$ indicates the position and $i$ indicates the dimension. That is, each dimension of the positional encoding corresponds to a sinusoid function, and the wavelengths form a geometric progression from $2\pi$ to $20000\pi$. The reason of choosing these functions is that they have been assumed theoretically in helping the model to learn to attend by relative positions easily, due to the characters that for any fixed offset $k$, $P E_{p o s}+k$ can be represented as a linear function of $P E_{p o s}$.

(2) Multi-head Self-Attention

Self-Attention is the major innovation in Transformer. But in implementation, rather than just computing the Self-Attention once, the multi-head mechanism runs through the scaled dot-product attention multiple times in parallel, and the outputs of these independent attention are then concatenated and linearly transformed into the expected dimensions. This multiple times Self-Attention computation is called Multi-head Self-Attention, which is applied for allowing the model to jointly attend to information from different representation sub-spaces at different positions.

(3) Encoder \& Decoder Blocks

The encoder is built by 6 identical components, each of which contains one multi-head attention layer with a fully connected network above it. These two sub-layers are equipped with residual connection as well as layer normalization. All the sub-layers have the same dimension of 512 in output data.

The decoder, however, is more complicated. There are also 6 components stacked; and in each component, three sub-layers are connected, including two sub-layers of multi-head self attention and one sub-layer of fully-connected neural network. Specifically, the bottom attention layer is modified with method called masked to prevent positions from attending to subsequent positions, which is used for avoiding the model to look into the future of the target sequence when predicting the current word. Additionally, the second attention layer (the top attention layer) performs multi-head attention over the output of the encoder stack.

\subsubsection{Transformer based NMT variants}
Due to the tremendous performance improvement by Transformer, related refinement has got huge attention for researchers. The well accepted weakness of vanilla Transformer includes: lacking of recurrence modeling, theoretically not Turing-complete, capturing position information, as well as large model complexity. All these drawbacks have hindered its further improvement of translation performance. In response to these problems, some adjustments have been proposed for getting a better model.

In respect of model architecture, some proposed modifications focused on both in depth of attention layer and network composition. Bapna et al. has proposed 2-3x deeper Transformer with a refined attention mechanism, which can be easier for the optimization of
deeper models \cite{Bapna2018}. The refined attention mechanism extended its connection to each encoder layers, like a weighted residual connections along the encoder depth, which allows the model to flexibly adjust the gradient flow to different layers of encoder. Similarly, Wang et al. \cite{Wang2019} proposed a more deeper Transformer model (25 layers of encoder), which continues the same line of Bapna et al.(2018)'s work with properly applying layer normalization and a novel output combination method. 

In contrast to the fixed layers of NMT model, Dehghani et al. proposed Universal Transformers, which cancelled to stack the constant number of layers by combining recurrent inductive bias of RNNs and Adaptive Computation Time halting mechanism, thus enhanced the original self-attention based representation for better learning iterative or recursive transformations. Notably, this adjustment has made the model be shown to be Turing-complete under certain assumptions\cite{Dehghani2018}.

As for refinement in network composition, inspired by the thinking of AutoML, So et al. applied neural architecture search (NAS) to find a comparable model with simplified architecture\cite{So2019}.  The Evolved Transformer proposed in \cite{So2019} has an innovative combination of basic blocks achieves the same quality as the original Transformer-Big model with 37.6\% less parameters. 

While most modification is focus on changing model structure directly, some new literature has chosen to utilize different input representation to improve model performance.  One direct method is using enhanced Position Encoding for sequence order injection, where vanilla Transformer has weakness in capturing position information. Shaw et al. proposed modified self-attention mechanism with awareness of utilizing representations of relative positions, which demonstrated to have a significant improvements in two MT tasks\cite{Shaw2018}.

Concurrently, using pre-initialized input representation with fine tune is another orientation, where some attempt have been proposed in different NLP tasks such as applying ELMo\cite{Peters2018} for encoder of NMT model\cite{Edunov2019}.  In terms of Transformer, one by-product of this innovative model is using self-attention for representing sequence, which can effectively fused word information with contextual information.  In later, Two well-known Transformer based input representation methods were been proposed named Bert (Bidirectional Encoder Representation from Transformers)\cite{Devlin2018} and GPT (Generative Pre-trained Transformer)\cite{Radford2018}, which has been indicated to bring improvement in some downstream NLP tasks. As for applying Transformer based pre-trained model in NMT task, more recently, this kind of trying has also been realized by using Bert as additional embedding layer or applying Bert as pre-trained model directly\cite{Clinchant2019}, which has been indicated to provide a bit better performance than vanilla Transformer after fine tune. Additionally, directly applying Bert as pre-trained model has been proved to have similar performance and thus can be more convenient for encoder initialization. 

\begin{figure}[h]
\centering
\includegraphics[width=18pc]{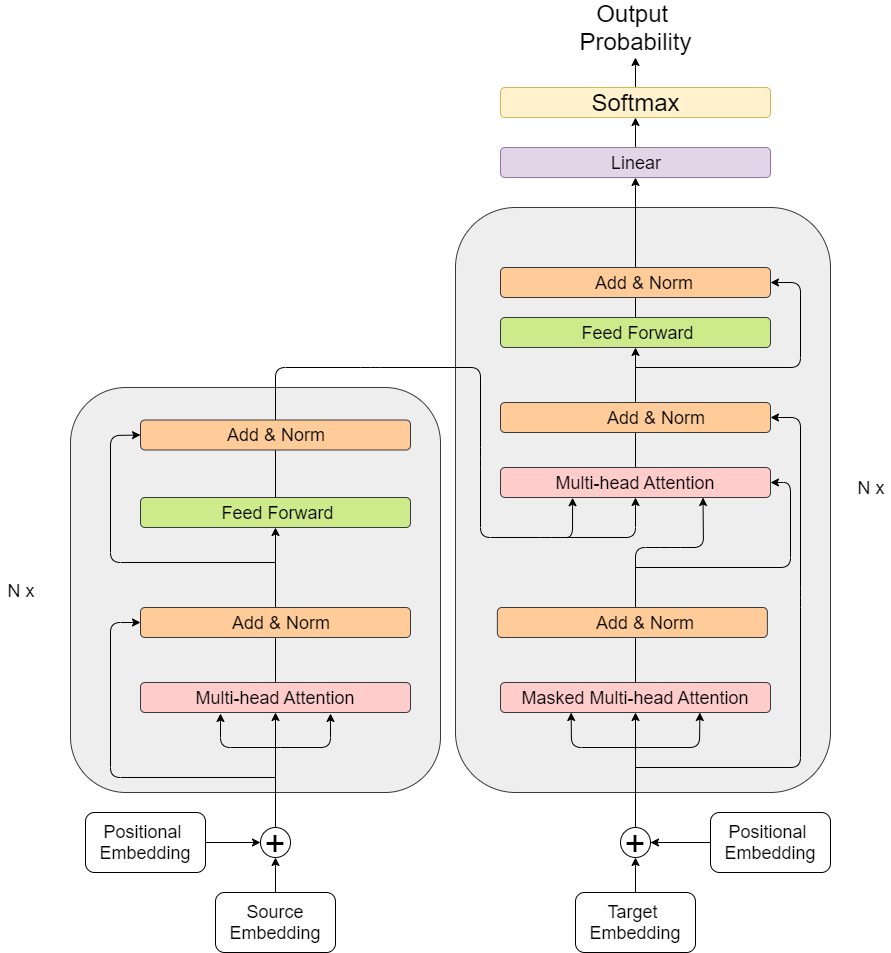}
\caption{The full structure of Transformer}
%
\label{fig_env14}
\end{figure}

The full structure of Transformer is illustrated in Fig.~\ref{fig_env14}.

\section{Future trend}
\label{sec:Sec7}
Although we have witnessed the fast-growing research progresses in NMT, there are still many challenges. Based on the extensive investigation \cite{Koehn2017}\cite{Sennrich2019}\cite{Galassi2019}\cite{Chaudhari2019}, we summarize the major challenges and list some potential directions in the following several aspects.

(1) In terms of translation performance, NMT still doesn't perform well in translating long sentences. This is mainly because of two reasons: the practical limitation in engineering and the learning ability of the model itself. For the first reason, some academic experiments have chosen to ignore part of the long sentence that exceeds the RNN length. But we do believe that it's not the same thing when NMT has been deployed in industrial applications. For the second reason, as research progresses, the model architecture would be more complicated. For example, Transformer model has applied innovative structure in its design which brought significant improvement in translation quality and speed\cite{Vaswani2017}. We believe that more refinements in model structure would be proposed. As we all know that RNN based NMT takes its advantages in modeling sequence order but results in computational inefficiency. More future work would consider the trade off between these two aspects.

(2) Alignment mechanism is essential for both SMT and NMT models. For the vast majority of NMT models, Attention Mechanism plays the functional role in the alignment task, while it arguably does broader work than conventional alignment models. We believe this advanced alignment method would still get attraction in future research, since powerful attention method can improve the model performance directly. Later research in attention mechanism would try to relieve the weakness in NMT such as interpretation ability\cite{Chaudhari2019}.

(3) Vocabulary coverage problem has always affected most of the NMT models. The research trend in handling computation load of softmax operation would pursue. And we have also found new training strategy which supports large vocabulary size. Besides, research of NMT operating in sub-word or character level has also aroused in recent years, which provided additional solution beyond traditional scope. More importantly, solving sophisticated translation scenario such as informal spelling is also a hot spot. Current NMT model integrated with character-level network has alleviated this phenomenon. Future work should focus on handling all kinds of OOV words in a more flexible way. 

(4) Low-Resource Neural Machine Translation\cite{Sennrich2019} is another hot spot in current NMT, which tries to solve the severe performance reduction when NMT model is trained with rare bilingual corpus. Since the aforementioned scenario happened commonly in practice where some seldom-used languages don't have enough data, we do believe this filed would be extended in further research. Multilingual translation method\cite{Firat2016}\cite{Dong2016} is the commonly proposed method which incorporated multi-language pair of data to improve NMT performance. It may need more interpretation about the different results in choosing different language pairs. Besides, unsupervised method has utilized the additional dataset and provided pre-trained model. Further research could improve its effect and provide hybrid training strategy with traditional method \cite{He2016}\cite{Ramachandran2016}\cite{Artetxe2017}.

(5) Finally, research in NMT applications would also become more abundant. Currently, many applications have been developed such as speech translation\cite{Duong2016}\cite{Weiss2017} and document level translation\cite{Wang2017b}. We believe that various applications (especially end-to-end tasks) would emerge in the future. We strongly hope that an AI based simultaneous translation system could be applied in large-scale, which can bring huge benefit to our human society\cite{Gu2016b}. 



\end{document}